\documentclass[conference]{IEEEtran}
\IEEEoverridecommandlockouts
\usepackage{cite}
\usepackage{amsmath,amssymb,amsfonts}
\usepackage{algorithmic}
\usepackage{graphicx}
\usepackage{textcomp}
\usepackage{xcolor}

\usepackage{cite}
\usepackage{url}
\usepackage{textcomp}

\usepackage{array}
\usepackage{multirow}
\usepackage{arydshln}

\usepackage{amsmath}
\usepackage{amssymb}
\usepackage{stfloats}
\usepackage{nicefrac}

\usepackage{graphicx}
\usepackage{tikz}
\usepackage{caption}
\usepackage{subcaption}
\usepackage{color}
\usepackage{tikz,pgfplots}
\usetikzlibrary{arrows}
\usetikzlibrary{calc}

\usepackage[acronym]{glossaries}

\usepackage{caption}

\usepackage{amsmath}
\usepackage{mathtools}
\usepackage{array}

\usepackage{physics}
\usepackage{amsmath}
\usepackage{tikz}
\usepackage{mathdots}
\usepackage{yhmath}
\usepackage{cancel}
\usepackage{color}
\usepackage{siunitx}
\usepackage{array}
\usepackage{multirow}
\usepackage{amssymb}
\usepackage{gensymb}
\usepackage{tabularx}
\usepackage{extarrows}
\usepackage{booktabs}
\usetikzlibrary{fadings}
\usetikzlibrary{patterns}
\usetikzlibrary{shadows.blur}
\usetikzlibrary{shapes}

\newcommand\blfootnote[1]{%
  \begingroup
  \renewcommand\thefootnote{}\footnote{#1}%
  \addtocounter{footnote}{-1}%
  \endgroup
}

\def\BibTeX{{\rm B\kern-.05em{\sc i\kern-.025em b}\kern-.08em
    T\kern-.1667em\lower.7ex\hbox{E}\kern-.125emX}}

\begin{document}

\title{SwiftLane: Towards Fast and Efficient Lane Detection\\

}


\author{\IEEEauthorblockN{
Oshada Jayasinghe, 
Damith Anhettigama, 
Sahan Hemachandra, 
Shenali Kariyawasam, \\ 
Ranga Rodrigo and 
Peshala Jayasekara}
\IEEEauthorblockA{Department of Electronic and Telecommunication Engineering,\\
University of Moratuwa, Sri Lanka\\}
\IEEEauthorblockN{Email: 
oshadajayasinghe@gmail.com,
damithkawshan@gmail.com,
sahanhemachandra@gmail.com, \\
shenali1997@gmail.com, 
ranga@uom.lk, peshala@uom.lk}}

\maketitle
\thispagestyle{plain}
\pagestyle{plain}

\def\footnoterule{\relax%
  \kern-5pt
  \hbox to \columnwidth{\vrule width .4\columnwidth height 0.4pt\hfill}
  \kern4.6pt}
\makeatother

\begin{abstract}
Recent work done on lane detection has been able to detect lanes accurately in complex scenarios, yet many fail to deliver real-time performance specifically with limited computational resources. In this work, we propose SwiftLane: a simple and light-weight, end-to-end deep learning based framework, coupled with the row-wise classification formulation for fast and efficient lane detection. This framework is supplemented with a false positive suppression algorithm and a curve fitting technique to further increase the accuracy. Our method achieves an inference speed of 411 frames per second, surpassing state-of-the-art in terms of speed while achieving comparable results in terms of accuracy on the popular CULane benchmark dataset. In addition, our proposed framework together with TensorRT optimization facilitates real-time lane detection on a Nvidia Jetson AGX Xavier as an embedded system while achieving a high inference speed of 56 frames per second. 
\end{abstract}

\blfootnote{\textcopyright \ 2021 IEEE. Personal use of this material is permitted.
  Permission from IEEE must be obtained for all other uses, in any current or future 
  media, including reprinting/republishing this material for advertising or promotional 
  purposes, creating new collective works, for resale or redistribution to servers or 
  lists, or reuse of any copyrighted component of this work in other works.},
 
\vspace{-1em}
  

\section{Introduction}

Lane detection is a pivotal element in driver assistance systems and autonomous vehicles as lane marker information is essential in maneuvering the vehicle safely on roads. Detecting lanes in real-world scenarios is a challenging task due to adverse weather, lighting conditions and occlusions. As the computational budget available for lane detection in the aforementioned systems is limited, a light-weight, fast and accurate lane detection system is crucial.

Recent lane detection approaches fall into two broad classes: semantic segmentation based methods and row-wise classification based methods. While semantic segmentation based methods \cite{pan2018SCNN,CurveLane-NAS,hou2019learning} provide competitive results in terms of accuracy, a common drawback is the reduced speed due to per-pixel classification and large backbones. On the other hand, row-wise classification based methods \cite{yoo2020end,qin2020ultra} focus on improving speed and obtaining real-time performance. However, the inherent limitation of a grid-based representation in row-wise classification methods and the bias towards overfitting due to the similar structure of lanes in the training set may result in reduced accuracy, highlighting the speed-accuracy trade-off in lane detection models.  

In this work, we propose a simple, light-weight, end-to-end deep learning based lane detection framework with a smaller backbone and a lesser number of multiply-accumulate operations (MACs) following the row-wise classification approach. The inference speed is significantly increased by reducing the computational complexity, and the light-weight network architecture is less prone to overfitting. Moreover, we also introduce a false positive suppression algorithm based on the length of the lane segment and the Pearson correlation coefficient, and a second-order polynomial fitting method as post-processing techniques to improve the overall accuracy of the system. Comprehensive experimental results are shown on the CULane \cite{pan2018SCNN} benchmark dataset, accompanied by a comparison of our results with other state-of-the-art approaches. An ablation study shows how each of the proposed methods contributes to the speed and the accuracy.

Furthermore, we deploy our lane detection framework on a Nvidia Jetson AGX Xavier integrated with Robot Operating System (ROS) \cite{ros} to demonstrate the capability of our light-weight network architecture to perform real-time lane detection in an embedded system. The trained model is optimized and quantized using TensorRT for increasing the inference speed. We also provide qualitative results for locally captured street view images to showcase how well our model generalizes for the task of lane detection.


In summary, our contributions are as follows: we introduce a novel, light-weight, end-to-end deep learning architecture supplemented with two effective post-processing techniques for fast and efficient lane detection. Our proposed method drastically improves the inference speed, reaching 411 frames per second (FPS) to surpass state-of-the-art while achieving comparable accuracy. We further optimize the trained model using TensorRT and implement it on an embedded system in the ROS ecosystem. The overall system achieves an inference speed of 56 FPS, demonstrating the capability of our method to perform real-time lane detection.

\section{Related Work}
\label{sec:related_works}
Initially, lane detection research mainly focused on classical image processing algorithms, such as using basic hand-crafted features\cite{kluge1995deform,yu1997lane,Ghazali2012road}, color-based approaches \cite{chiu2005lane,lee2009effective}, and traditional feature extraction methods with machine learning algorithms such as decision trees and support vector machines \cite{gonzalez2000lane,lanesvm2012}. Although these methods are computationally less expensive, the performance is poor in complex scenarios with occlusions, shadows and different lighting conditions.

Recent deep learning based approaches outperform classical methods and can be further divided into two broad classes: semantic segmentation based methods and row-wise classification based methods. In semantic segmentation based methods \cite{pan2018SCNN,CurveLane-NAS,hou2019learning}, classification is done on a per-pixel basis by classifying each pixel as lane or background. A special convolution method known as slice-by-slice convolution is proposed in SCNN \cite{pan2018SCNN}, which enables information propagation within the same layer to improve the detection of long thin structures such as lanes. CurveLane-NAS \cite{CurveLane-NAS} focuses on capturing long-range contextual information and short-range curved trajectory information using a lane-sensitive neural architecture search framework. Attention maps extracted from different layers of a trained model which contain important contextual information are used as distillation targets for the lower layers in SAD \cite{hou2019learning}. The pixel-wise computation in semantic segmentation based approaches increases the computational complexity and reduces the inference speed drastically.

Row-wise classification based methods \cite{yoo2020end, qin2020ultra} have been able to progress towards real-time lane detection by addressing the computational complexity problem. In these approaches, the input image is divided into a grid and for each row, the model outputs the probability of each cell belonging to a lane. This approach is first introduced in E2E-LMD \cite{yoo2020end} by converting the output of the segmentation backbone to a row-wise representation using a special module called horizontal reduction module. The no-visual-clue problem in lane detection is addressed in UltraFast \cite{qin2020ultra} using a low-cost, row-wise classification based network, which utilizes global and structural information. Although their approach achieves state-of-the-art speed of 322.5 FPS, the accuracy is low when compared with other methods.






\begin{figure*}
    \begin{center}
        \input{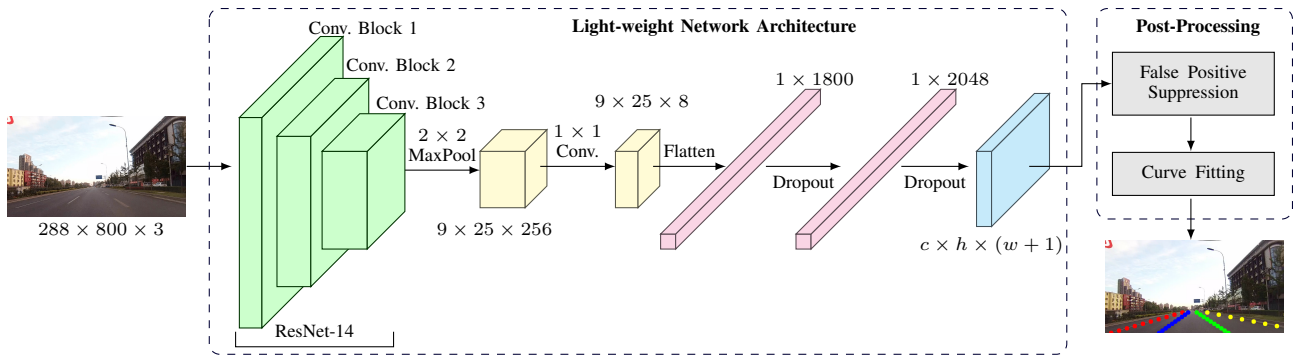}
        \caption{Proposed model architecture. ResNet-14 backbone generates feature maps from the input image. A $ 2 \times 2 $ max pooling layer and a $ 1 \times 1 $ convolutional layer are used to reduce the spatial dimensions and the number of channels. Resulting feature maps are flattened and passed through two fully-connected layers with dropout layers in between. The model predictions are fed through false positive suppression and curve fitting modules to obtain the lane output.}
        
        \label{fi:architecture}
    \end{center}
    \vspace{-2ex}
\end{figure*}

\begin{figure}
    \centering
    \input{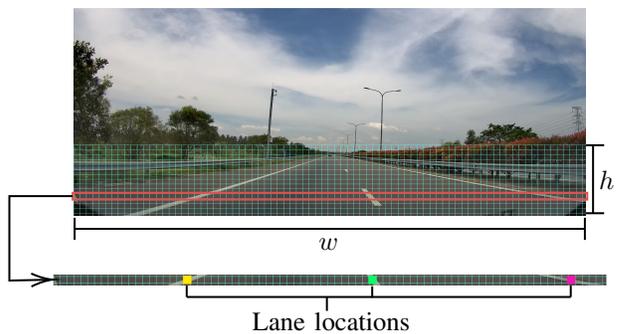}
    \caption{Lane Representation. The region comprising lanes is divided into a pre-defined number of row anchors ($h$) and gridding cells ($w$).}
    \vspace{-0.3cm}
    \label{fig:lane_rep}
\end{figure}

Almost all of the above mentioned algorithms have been implemented in high-end computational platforms and implementation of lane detectors in embedded systems is comparatively a less researched area. A lane detection algorithm optimized for PXA255 embedded device has been introduced by \cite{Ming07real} which achieves a frame rate of 13 FPS. PathMark \cite{pathmark} is another lane detection algorithm running at 13 FPS in a TI-OMAP4430 based embedded system. A Nvidia Jetson-TK1 board has been used in \cite{lanedeparture} for implementing a real-time lane detection and departure warning system at 44 FPS. In \cite{marcos18fast}, a lane detection and modeling pipeline has been presented for embedded platforms which delivers real-time performance in a Jetson-TX2 embedded device. All of these approaches rely on classical image processing based techniques and do not perform well in complex scenarios when compared with deep learning based approaches.


\section{Methodology}
\label{sec:method}
In this section, we present the lane representation mechanism, a detailed explanation of our model architecture and the algorithms used to further increase the model accuracy.

\subsection{Lane Representation}
\label{ssec:laneRep}

We address the lane detection task as a row-wise classification problem following the formulation introduced by \cite{qin2020ultra}. The region of the image which contains lanes is divided into a pre-defined number of row anchors ($h$) and each row anchor is divided into a pre-defined number of gridding cells ($w$) as shown in Fig. \ref{fig:lane_rep}. The number of lanes ($c$) is pre-defined, and for each lane, the lane locations are represented by a $h \times w$ grid. An additional cell is attached to the end of each row anchor to indicate the absence of a particular lane in that row anchor.

\subsection{Model Architecture}
\label{ssec:Arch}


We propose a simple end-to-end light-weight convolutional neural network based model architecture for the lane detection task as shown in Fig. \ref{fi:architecture}. The first stage of the proposed model is the backbone which extracts features from the input image. As the backbone we use ``ResNet-14'' which is obtained by dropping the last four convolutional layers of ResNet-18 \cite{he2016deep} to increase the speed by reducing the computational complexity. 

The output of the backbone is a feature representation of the image which would then be fed into a $2 \times 2$ max pooling layer for dimensionality reduction in the spatial dimensions. For dimensionality reduction in the channel dimension a $1 \times 1$ convolution layer is applied. This output is flattened to obtain a one-dimensional tensor which is then passed through two fully connected layers to obtain the output tensor. Dropout layers are implemented in between to further prevent the network from overfitting.

The output tensor represents the score of each gridding cell (including the no lane cell) belonging to each lane in each row anchor. $S_{i,j,k}$ represents the score of $k$\textsuperscript{th} gridding cell in $j$\textsuperscript{th} row anchor belonging to $i$\textsuperscript{th} lane which can be obtained by, 
\begin{equation}
\label{eq:lane_rep}
S_{i,j,k}=f(X), \;
\text{s.t.} \; i\in[1,c], \; j\in[1,h], \; k\in[1,w+1]
\end{equation}
\normalsize
Here, $f$, $X$, $c$, $h$ and $w$ stands for the classification model, the input image, the number of lanes, the number of row anchors and the number of gridding cells, respectively. The lane points can then be extracted by choosing the gridding cell with the highest score in each row anchor for each lane. If the last gridding cell is not the cell with the highest score, the location of $i$\textsuperscript{th} lane in $j$\textsuperscript{th} row anchor is given by, 

\begin{equation}
\label{eq:location}
Loc_{i,j}=\mathrm{argmax}_{k}\; (S_{i,j,k})\;, \; \text{s.t.} \; k \in [1,w] 
\end{equation}

Having the highest score in the last gridding cell implies that the considered lane is not present in the selected row anchor. For training the model, we define the classification loss as the negative log likelihood loss which is given by, 
\begin{equation}
\label{eq:classLoss}
L_{cls}= \sum_{i=1}^{C} \sum_{j=1}^{h} -\alpha_{i,j,T_{i,j}} \cdot \log( P_{i,j,T_{i,j}})
\end{equation}
Here, $T_{i,j}$ denotes the correct location (gridding cell) of $i$\textsuperscript{th} lane in $j$\textsuperscript{th} row anchor as per the ground truth and $P_{i,j,k}$ denotes the probability of $k$\textsuperscript{th} gridding cell in $j$\textsuperscript{th} row anchor belonging to $i$\textsuperscript{th} lane which can be obtained by,
\begin{equation}
\label{eq:prob}
P_{i,j,k}=\mathrm{softmax}(S_{i,j,k})
\end{equation}
 $\alpha_{i,j,k}$ is the modulating factor for the focal loss adjustment as mentioned in \cite{lin2017focal}.  
\begin{equation}
\label{eq:alpha}
\alpha_{i,j,k}=(1-P_{i,j,k})^{\gamma}
\end{equation}

\subsection{False Positive Suppression}
\label{ssec:FPSup}

We propose two post processing techniques to reduce false detections in the model output. First, we remove all instances of small lane segments which have a less number of detected lane points than a threshold value. Second, we remove all instances of lanes which have a considerable deviation from a straight line. Pearson correlation coefficient measures the linear correlation between two variables which is given by (\ref{eq:pearson}), where $x \textsubscript{i}$ and $y\textsubscript{i}$ are the sample data points of $x$ and $y$ variables and $\bar{x}$ and $\bar{y}$ are the respective means. 
\begin{equation}
\label{eq:pearson}
r= \frac{\sum_{}^{}(x\textsubscript{i}-\bar{x})(y\textsubscript{i} - \bar{y})}{\sqrt{\sum_{}^{}(x\textsubscript{i}-\bar{x})^{2}(y\textsubscript{i} - \bar{y})^{2}}}
\end{equation}

In our case, Pearson correlation coefficient of row anchors and gridding cells of an identified lane segment is used to measure how well the lane points can be represented using a straight line. Since majority of the lanes have a slight deviation from a straight line, the Pearson correlation coefficient should be close to one in magnitude. Therefore, we remove all instances of lanes which have a Pearson correlation coefficient below a threshold value.

\subsection{Curve Fitting}
\label{ssec:LSQLF}
In most of the scenarios, lanes are straight lines or curve segments with small curvature values. Therefore, lanes can be approximated to a greater extent by second-order polynomials. Since we use a finite number of gridding cells, lanes in the model output are represented in the discrete domain. Second-order polynomial fitting can be used to replace these discrete gridding cell numbers by continuous values which results in smooth lane segments.

\section{Experiments}
\label{sec:Experiments}
In this section, we present the details about the dataset used to evaluate our model, the training process and a detailed description on the embedded system implementation for real-time applications.

\subsection{Dataset Description}
\label{ssec:datasets}

For the training and quantitative evaluation of our model, we use the publicly available CULane \cite{pan2018SCNN} benchmark dataset which is one of the largest lane detection datasets with 133,235 total frames having a resolution of $1640 \times 590$. The dataset is divided into the train set, the validation set and the test set which comprises 88,880 frames, 9,675 frames and 34,680 frames, respectively. The dataset covers several complex scenarios and the test images are divided into 9 categories: Normal, Crowded, Dazzle light, Shadow, No line, Arrow, Curve, Crossroad and Night. 

As the evaluation metric, F1-measure is used to compare the performance in the CULane benchmark. Each lane is represented by a 30-pixel-width line and each prediction which has an intersection over union (IoU) greater than 0.5 with the ground truth is considered as a true positive. Then F1-measure is calculated as follows where $TP$, $FP$ and $FN$ stands for true positives, false positives and false negatives, respectively.

\begin{equation}
precision = \frac{TP}{TP+FP}
\end{equation}

\begin{equation}
recall = \frac{TP}{TP+FN}
\end{equation}

\begin{equation}
    F_{1}-measure = \frac{2\times precision \times recall}{precision+recall}
\end{equation}

\begin{table*}[t]
\addtolength{\tabcolsep}{-2.5pt}
\begin{center}
\caption{Comparison of F1-measure and speed (FPS) on CULane with state-of-the-art methods}
\label{ta:results}

 \normalsize 
\begin{tabular}{|@{}l|c|c|c|c|c|c|c|c|c|c|c|@{}c|}
 \hline
 \textbf{Model}                     & \textbf{Normal} & \textbf{Crowd} & \textbf{Dazzle} & \textbf{Shadow} & \textbf{No Line} & \textbf{Arrow} & \textbf{Curve} & \textbf{Cross} & \textbf{Night} & \textbf{Total} & \textbf{FPS} & \textbf{GMACs}\\ 
 \hline
 SCNN\cite{pan2018SCNN}             & 90.6          & 69.7          & 58.5          & 66.9          & 43.4          & 84.1          & 64.4          & 1990          & 66.1          & 71.6          & 7.5   & - \\
 \hline
  ENet-SAD\cite{hou2019learning}    & 90.1          & 68.8          & 60.2          & 65.9          & 41.6          & 84.0          & 65.7          & 1998          & 66.0          & 70.8          & 75    & -\\
  \hline
 ERFNet-E2E \cite{yoo2020end}       & 91.0          & 73.1          & 64.5          & 74.1          & 46.6          & 85.8          & 71.9          & 2022          & 67.9          & 74.0          & -     & -\\
 \hline
 CurveLane-S\cite{CurveLane-NAS}    & 88.3          & 68.6          & 63.2          & 68.0          & 47.9          & 82.5          & 66.0          & 2817          & 66.2          & 71.4          & -     & 9.0 \\
 CurveLane-M\cite{CurveLane-NAS}    & 90.2          & 70.5          & 65.9          & 69.3          & 48.8          & 85.7          & 67.5          & 2359          & 68.2          & 73.5          & -     & 33.7 \\
 CurveLane-L\cite{CurveLane-NAS}    & 90.7          & 72.3          & 67.7          & 70.1          & 49.4          & 85.8          & 68.4          & 1746          & 68.9          & 74.8          & -     & 86.5 \\
 \hline
  PINet \cite{pinet_2021}             & 90.3          & 72.3          & 66.3          & 68.4          & 49.8          & 83.7          & 65.6          & 1427          & 67.7          & 74.4          & 25    & - \\
 \hline
 UltraFast-18\cite{qin2020ultra}    & 87.7          & 66.0          & 58.4          & 62.8          & 40.2          & 81.0          & 57.9          & 1743          & 62.1          & 68.4          & 361   & 8.4\\
 UltraFast-34\cite{qin2020ultra}    & 90.7          & 70.2          & 59.5          & 69.3          & 44.4          & 85.7          & \textbf{69.5} & 2037          & 66.7          & 72.3          & 217   & 16.9 \\
 \hline
 RESA-34 \cite{resa_2020}         & 91.9          & 72.4          & 66.5          & 72.0          & 46.3          & 88.1          & 68.6          & 1896          & 69.8          & 74.5          & 45.5  & - \\
 RESA-50 \cite{resa_2020}         & 92.1          & 73.1          & 69.2          & 72.8          & 47.7          & 88.3          & 70.3          & 1503          & 69.9          & 75.3          & 35.7  & - \\
\hline
 LaneATT-18\cite{tabelini2021keep}  & 91.2          & 72.7          & 65.8          & 68.0          & 49.1          & 87.8          & 63.8          & \textbf{1020} & 68.6          & 75.1          & 250   & 9.3 \\
 LaneATT-34\cite{tabelini2021keep}  & 92.1          & 75.0          & 66.5          & 78.2          & 49.4          & 88.4          & 67.7          & 1330          & 70.7          & 76.7          & 171   & 18.0\\
 LaneATT-122\cite{tabelini2021keep} & 91.7          & 76.2          & 69.5          & 76.3          & 50.5          & 86.3          & 64.1          & 1264          & 70.8          & 77.0          & 26    & 70.5\\
 \hline
FOLOLane \cite{FOLO_2021_CVPR}    & \textbf{92.7} & \textbf{77.8} & \textbf{75.2} & \textbf{79.3} & \textbf{52.1} & \textbf{89.0} & 69.4          & 1569          & \textbf{74.5} & \textbf{78.8} & 40    & -\\
\hline
 SwiftLane (Ours)                   & 90.46         & 71.07         & 62.51         & 73.69         & 46.17         & 85.00         & 64.92         & 1096          & 68.77         & 74.03         & \textbf{411} & \textbf{6.52}\\
 \hline

\end{tabular}

\vspace{-2ex}
\end{center}
\end{table*}

\subsection{Model Training}
\label{ssec:imp_details}

Each image in the CULane dataset is resized to $288 \times 800$ from the input resolution of $590 \times 1640$. We use 36 row anchors ($h$) and 150 gridding cells ($w$) to represent the area which contains lanes (height ranging from 260 to 590 in the original image). The number of lanes ($c$) is set to 4. The threshold for false positive suppression using number of lane points is set to 12, and the threshold for false positive suppression using the Pearson correlation coefficient is set to 0.995.

As the optimization algorithm, SGD with momentum \cite{pmlr-v28-sutskever13} is used with an initial learning rate of 0.1, a momentum of 0.9 and a weight decay of $1 \times 10^{-4}$ for training the model. The model is trained for 50 epochs and at 15\textsuperscript{th}, 25\textsuperscript{th}, 35\textsuperscript{th} and 45\textsuperscript{th} epochs, the learning rate is multiplied by a factor of 0.3. For training and testing our model we use a computational platform comprising of an Intel Core i9-9900K CPU and Nvidia RTX-2080 Ti GPU. All experiments are carried out using PyTorch\cite{paszke2017automatic} based on the implementation of \cite{qin2020ultra}.

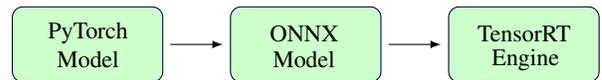
\begin{figure}[t]
    \begin{center}
        \begin{tikzpicture}

\setlength{\baselineskip}{5em}

\tikzstyle{bluebox}=[draw=cyan!20!black,fill=cyan!20]
\tikzstyle{greenbox}=[draw=green!20!black,fill=green!20]
\tikzstyle{orangebox}=[draw=yellow!20!black,fill=yellow!20]
\tikzstyle{magentabox}=[draw=magenta!20!black,fill=magenta!20]

\tikzstyle{labelnode}=[align=center, execute at begin node=\setlength{\baselineskip}{0.75em}]

\newcommand{\cube}[5][]
{
	\pgfmathsetmacro{\cubex}{(#2)}
	\pgfmathsetmacro{\cubey}{{#3}}
	\pgfmathsetmacro{\cubez}{{#4}}
	\draw[#5] (0,0,0) -- ++(-\cubex,0,0) -- ++(0,-\cubey,0) -- ++(\cubex,0,0) -- cycle;
	\draw[#5] (0,0,0) -- ++(0,0,-\cubez) -- ++(0,-\cubey,0) -- ++(0,0,\cubez) -- cycle;
	\draw[#5] (0,0,0) -- ++(-\cubex,0,0) -- ++(0,0,-\cubez) -- ++(\cubex,0,0) -- cycle;
}
\begin{scope}[xshift=.3cm, yshift=-.9cm]
\draw [draw=blue!20!black,fill=green!20, rounded corners] (0,0) rectangle ++(2,1);
\node at (0.35, .65) [anchor=west] {\small PyTorch};
\node at (0.47, .3) [anchor=west] {\small Model};
\draw [-latex] (2.1, .5) -- ++(.7, 0);
\draw [draw=blue!20!black,fill=green!20, rounded corners] (2.9,0) rectangle ++(2,1);
\node at (3.3, .65) [anchor=west] {\small ONNX};
\node at (3.34, .3) [anchor=west] {\small Model};
\draw [-latex] (5, 0.5) -- ++(.7, 0);
\draw [draw=blue!20!black,fill=green!20, rounded corners] (5.8,0) rectangle ++(2,1);
\node at (6.06, .65) [anchor=west] {\small TensorRT};
\node at (6.25, .3) [anchor=west] {\small Engine};
\end{scope}

\end{tikzpicture}
        \vspace{1ex}
        \caption{Optimization of the lane detection model. The trained PyTorch model is converted to a TensorRT engine.}
        
        \label{fi:oppt}
    \end{center}
    \vspace{-0.5em}
\end{figure}

As a measure to make the model more robust and generalized without overfitting, we apply two data augmentation techniques while training the model. First, we fit a random affine transformation to each image which comprises a random rotation, a random horizontal shift and a random vertical shift. Second, we use the colour jitter augmentation technique to randomly change the brightness and the contrast of the input image.

\subsection{Implementation on the Embedded System}
\label{ssec:embedded}

As the embedded system, we use a Nvidia Jetson AGX Xavier which possesses the required processing power to run deep learning based algorithms with the help of CUDA and Tensor cores. We further optimize our lane detection model for the embedded system by generating a TensorRT engine as shown in Fig. \ref{fi:oppt}. First, the trained PyTorch model is converted to ONNX file format and the ONNX model is then used by the ONNXParser in TensorRT Python API to generate the TensorRT engine. We evaluate the use of both single-precision floating point (FP32) and half-precision floating point (FP16) formats for building the TensorRT engine.

\begin{figure}[t]
    \centering
\includegraphics[width=0.98\linewidth]{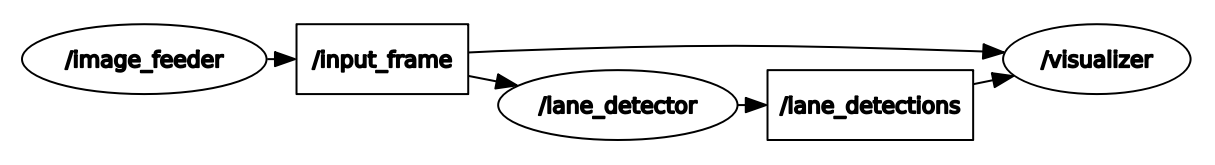}

    \caption{RQT graph for the implementation of the lane detection system in the ROS ecosystem.}
    \vspace{-0.3cm}
    \label{fig:rqt_graph}
\end{figure}

We implement the lane detector system in the Robot Operating System (ROS) \cite{ros} ecosystem as shown in Fig. \ref{fig:rqt_graph}.  The \textit{image\_feeder} node retrieves frames from a given video file and publishes each frame to the \textit{input\_frame} topic. The \textit{lane\_detector} node detects lanes in the current frame and publishes the detections to the \textit{lane\_detections} topic. For a faster inference speed, we use the FP16 quantized TensorRT engine for the lane detection task. The visualizer node marks the detected lane points in the current frame and publishes the resultant image to the \textit{output\_frame} topic. The RViz visualization tool is used to visualize the lane detections in real-time.


\section{Results}
\label{sec:results}

\normalsize

\begin{figure*}
     \centering
     \begin{subfigure}[b]{0.24\linewidth}
         \centering
         \includegraphics[width=0.98\linewidth]{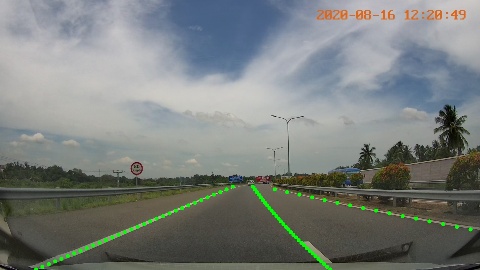}
     \end{subfigure}%
     \begin{subfigure}[b]{0.24\linewidth}
         \centering
         \includegraphics[width=0.98\linewidth]{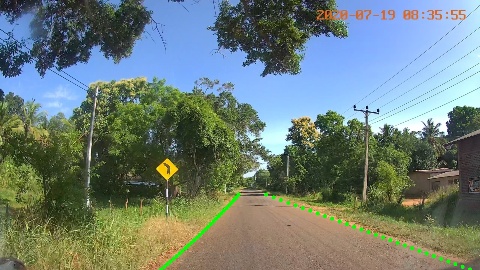}
     \end{subfigure}%
     \begin{subfigure}[b]{0.24\linewidth}
         \centering
         \includegraphics[width=0.98\linewidth]{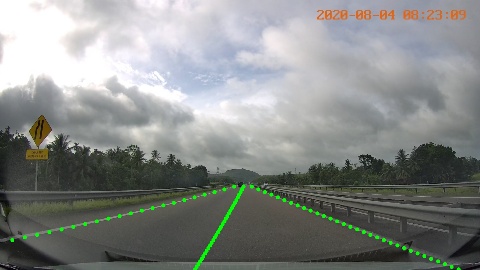}
     \end{subfigure}%
          \begin{subfigure}[b]{0.24\linewidth}
         \centering
         \includegraphics[width=0.98\linewidth]{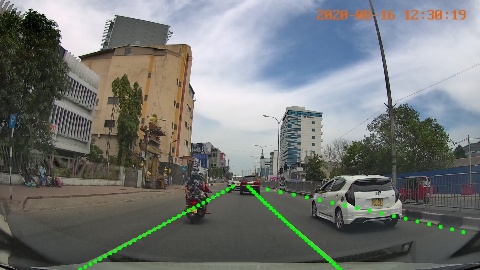}
     \end{subfigure}%
     \vspace{0.5em}
     
         \begin{subfigure}[b]{0.24\linewidth}
         \centering
         \includegraphics[width=0.98\linewidth]{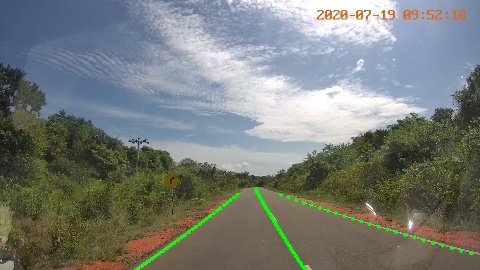}
     \end{subfigure}%
     \begin{subfigure}[b]{0.24\linewidth}
         \centering
         \includegraphics[width=0.98\linewidth]{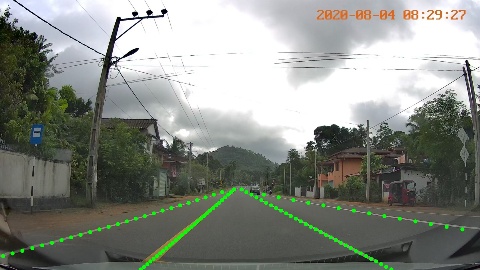}
     \end{subfigure}%
     \begin{subfigure}[b]{0.24\linewidth}
         \centering
         \includegraphics[width=0.98\linewidth]{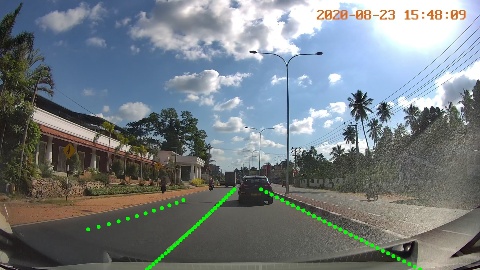}
     \end{subfigure}%
          \begin{subfigure}[b]{0.24\linewidth}
         \centering
         \includegraphics[width=0.98\linewidth]{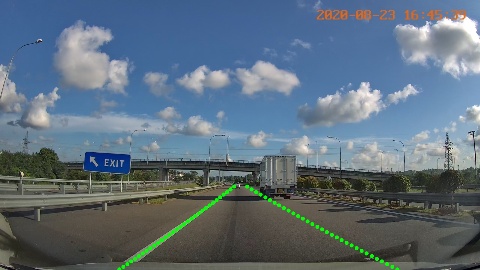}
     \end{subfigure}%
     \vspace{0.5em}
     
     \caption{ 
    Visualization of lane detection result on locally captured images. The first six images show accurate detections while the last two show failure cases including false detections and undetected lanes.}
     \label{fi:local_lane_qualitative_results}
\end{figure*}

\begin{figure}
     \centering
     \begin{subfigure}[b]{0.32\linewidth}
         \centering
         \includegraphics[width=0.98\linewidth]{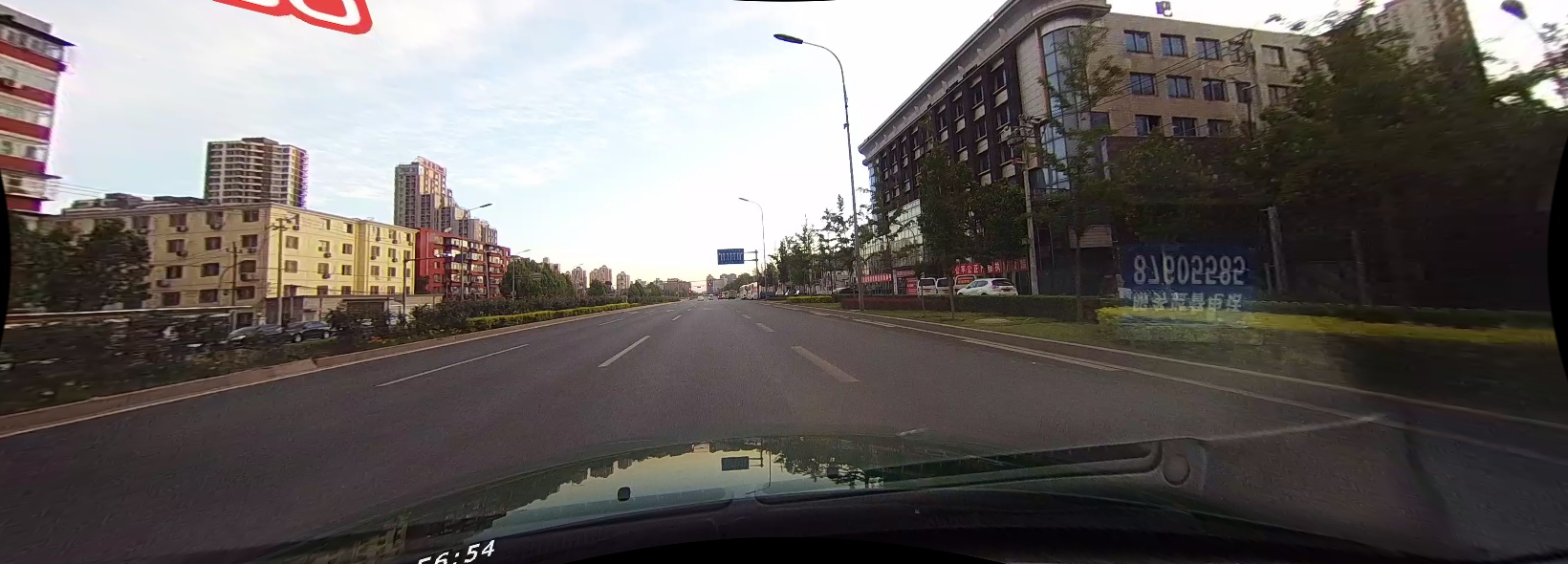}
     \end{subfigure}%
     \begin{subfigure}[b]{0.32\linewidth}
         \centering
         \includegraphics[width=0.98\linewidth]{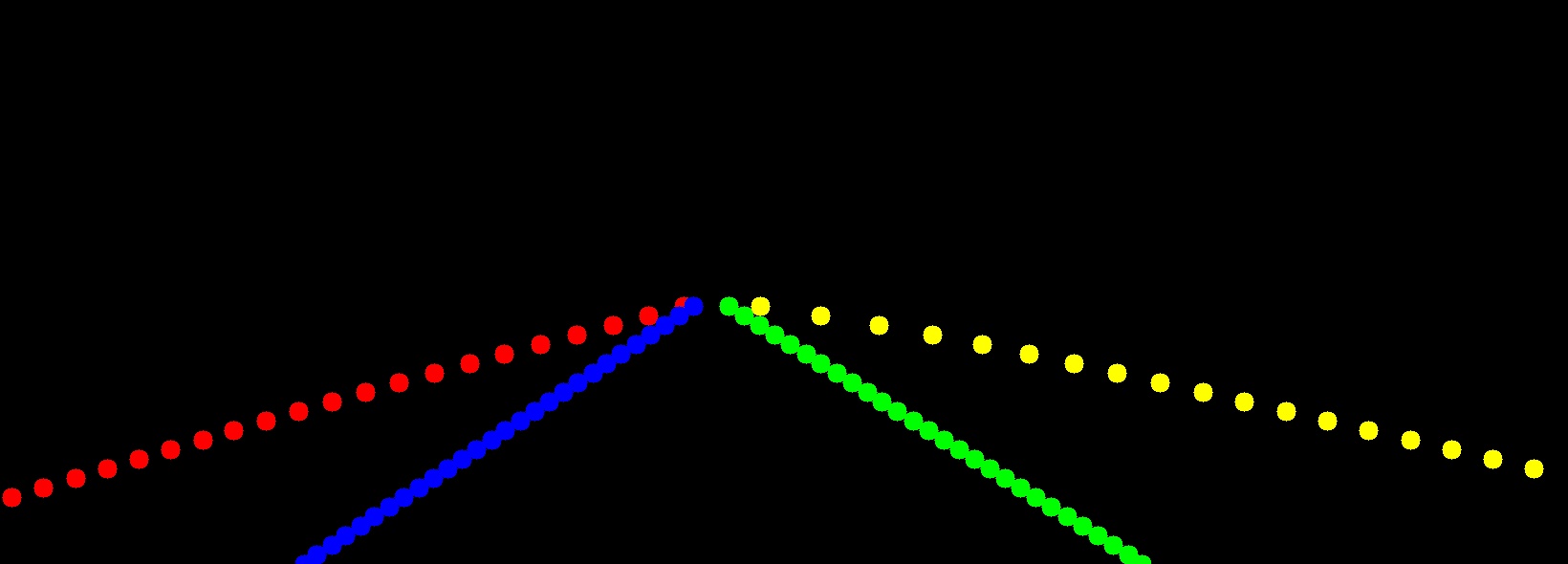}
     \end{subfigure}%
     \begin{subfigure}[b]{0.32\linewidth}
         \centering
         \includegraphics[width=0.98\linewidth]{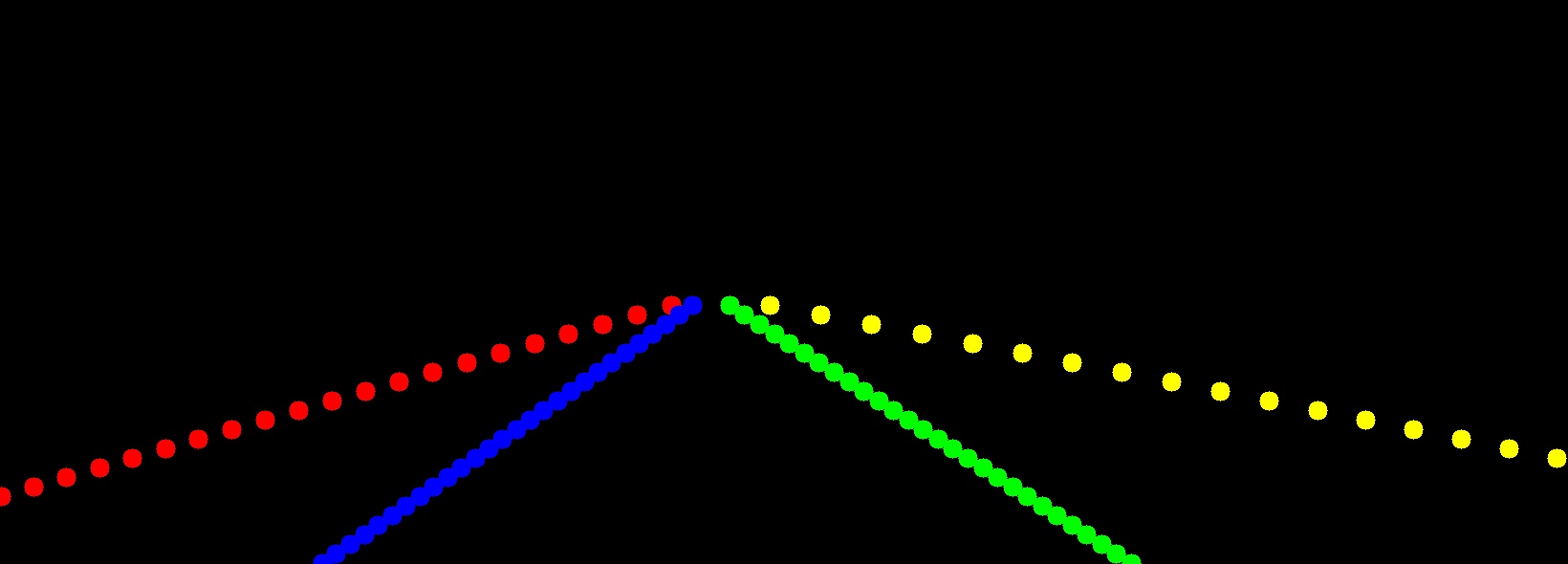}
     \end{subfigure}%
     \vspace{.3ex}
     
     \begin{subfigure}[b]{0.32\linewidth}
         \centering
         \includegraphics[width=0.98\linewidth]{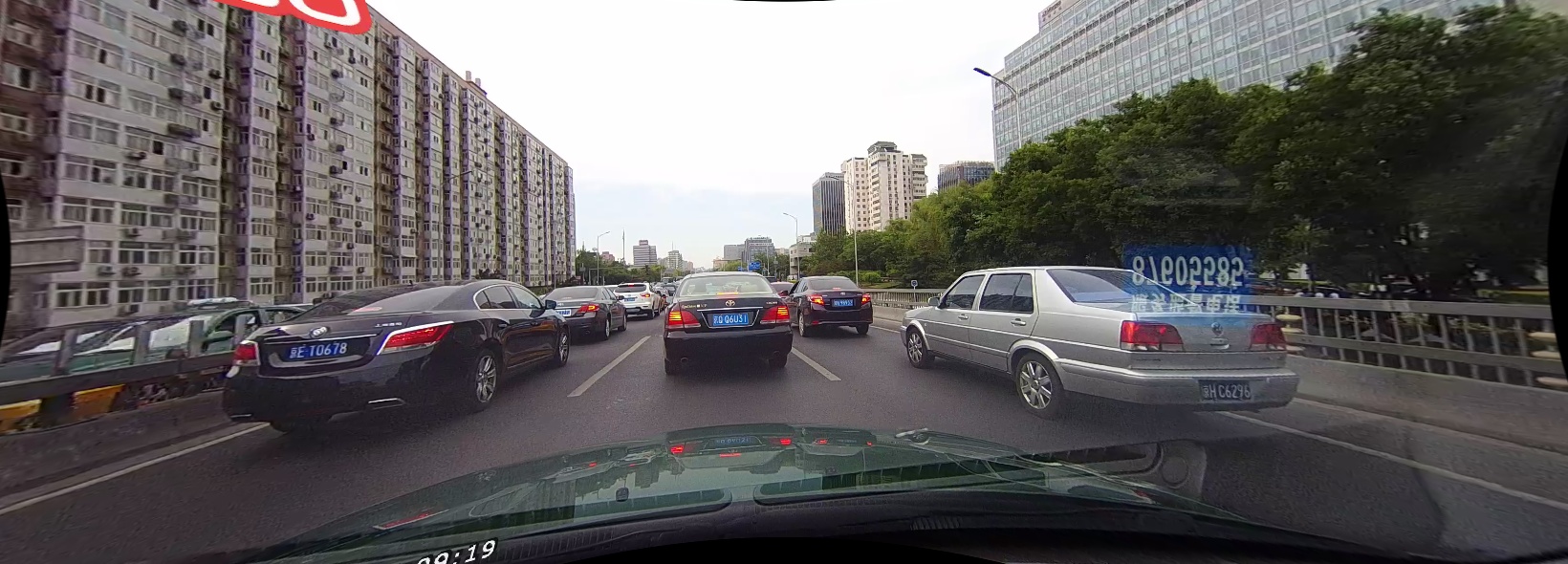}
     \end{subfigure}%
     \begin{subfigure}[b]{0.32\linewidth}
         \centering
         \includegraphics[width=0.98\linewidth]{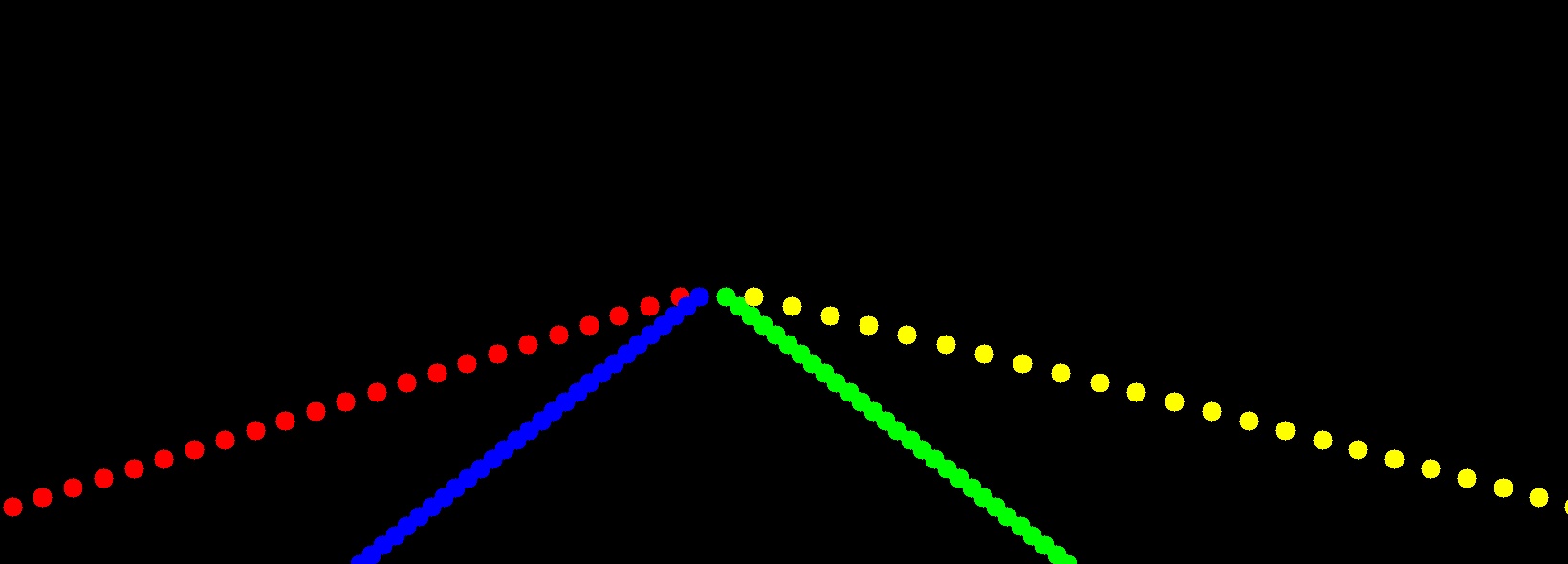}
     \end{subfigure}%
     \begin{subfigure}[b]{0.32\linewidth}
         \centering
         \includegraphics[width=0.98\linewidth]{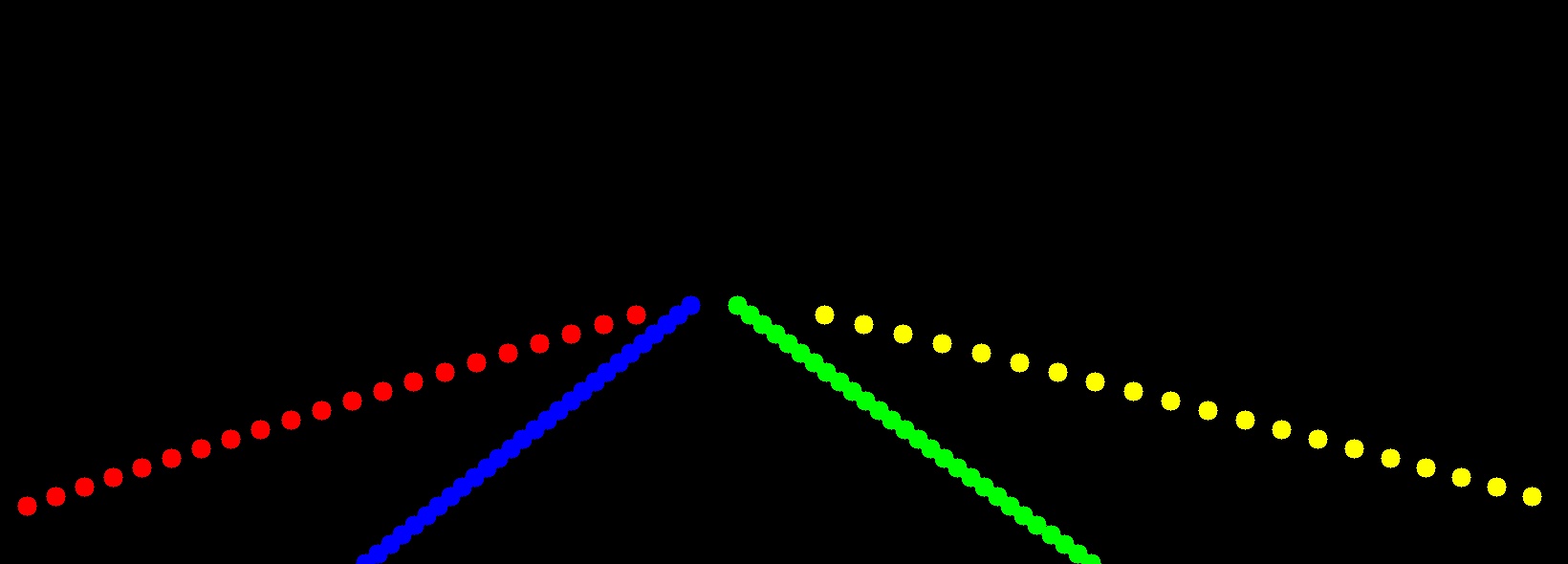}
     \end{subfigure}%
     \vspace{.3ex}
     
     \begin{subfigure}[b]{0.32\linewidth}
         \centering
         \includegraphics[width=0.98\linewidth]{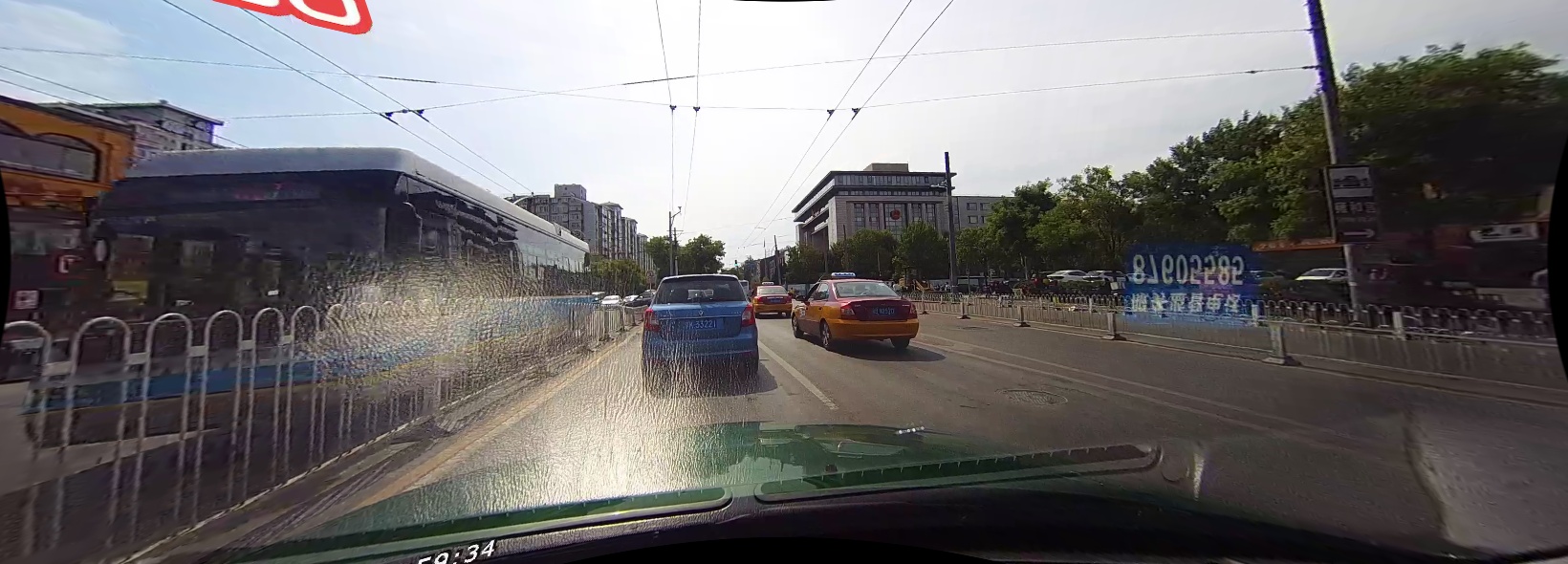}
     \end{subfigure}%
     \begin{subfigure}[b]{0.32\linewidth}
         \centering
         \includegraphics[width=0.98\linewidth]{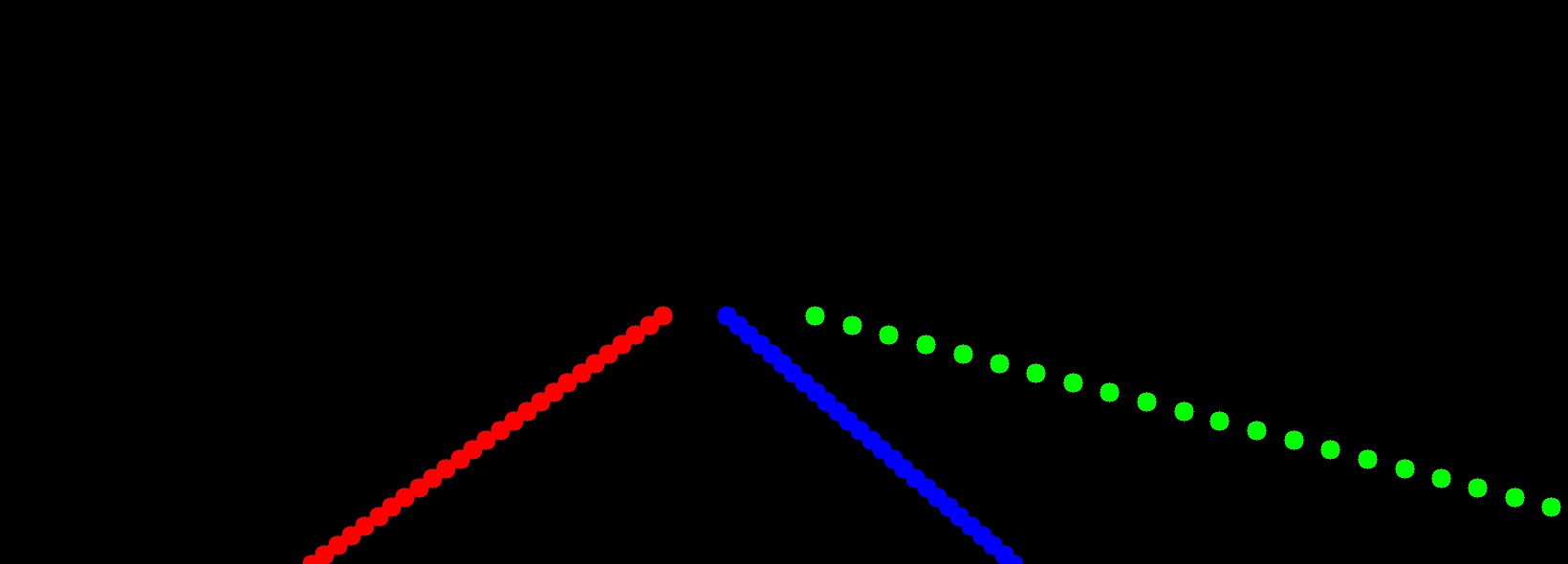}
     \end{subfigure}%
     \begin{subfigure}[b]{0.32\linewidth}
         \centering
         \includegraphics[width=0.98\linewidth]{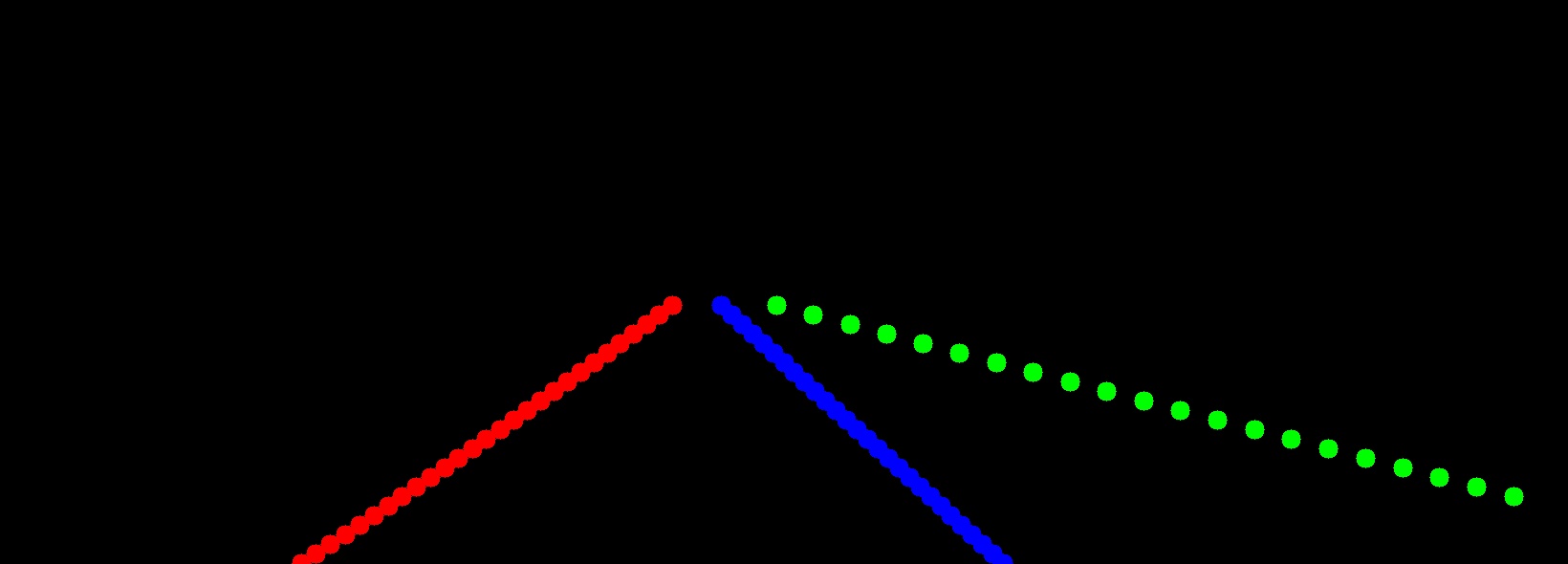}
     \end{subfigure}%
     \vspace{.3ex}
     
     \begin{subfigure}[b]{0.32\linewidth}
         \centering
         \includegraphics[width=0.98\linewidth]{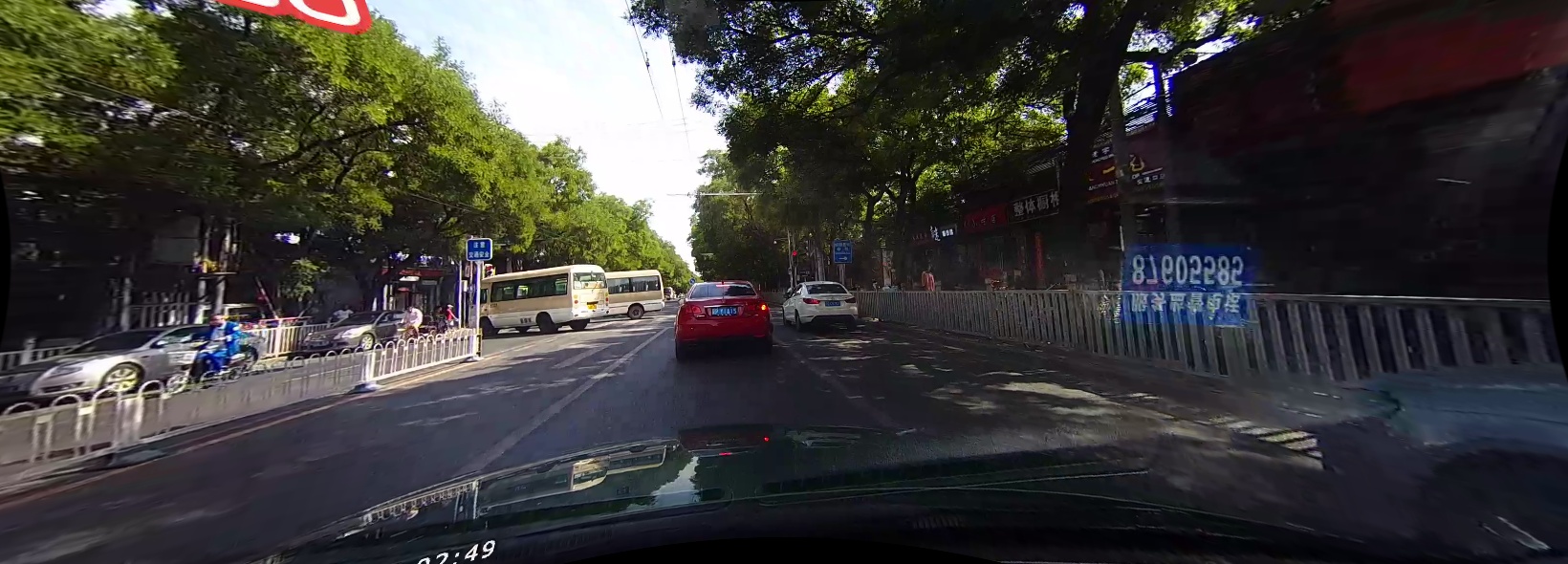}
     \end{subfigure}%
     \begin{subfigure}[b]{0.32\linewidth}
         \centering
         \includegraphics[width=0.98\linewidth]{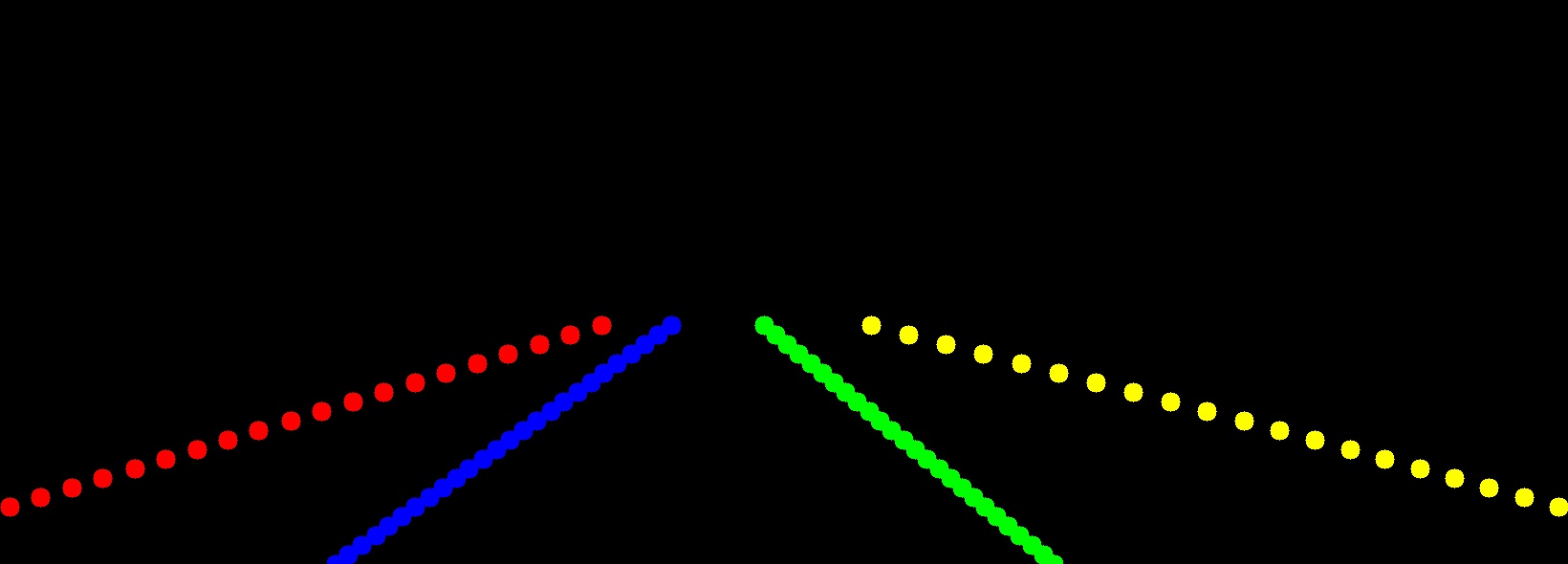}
     \end{subfigure}%
     \begin{subfigure}[b]{0.32\linewidth}
         \centering
         \includegraphics[width=0.98\linewidth]{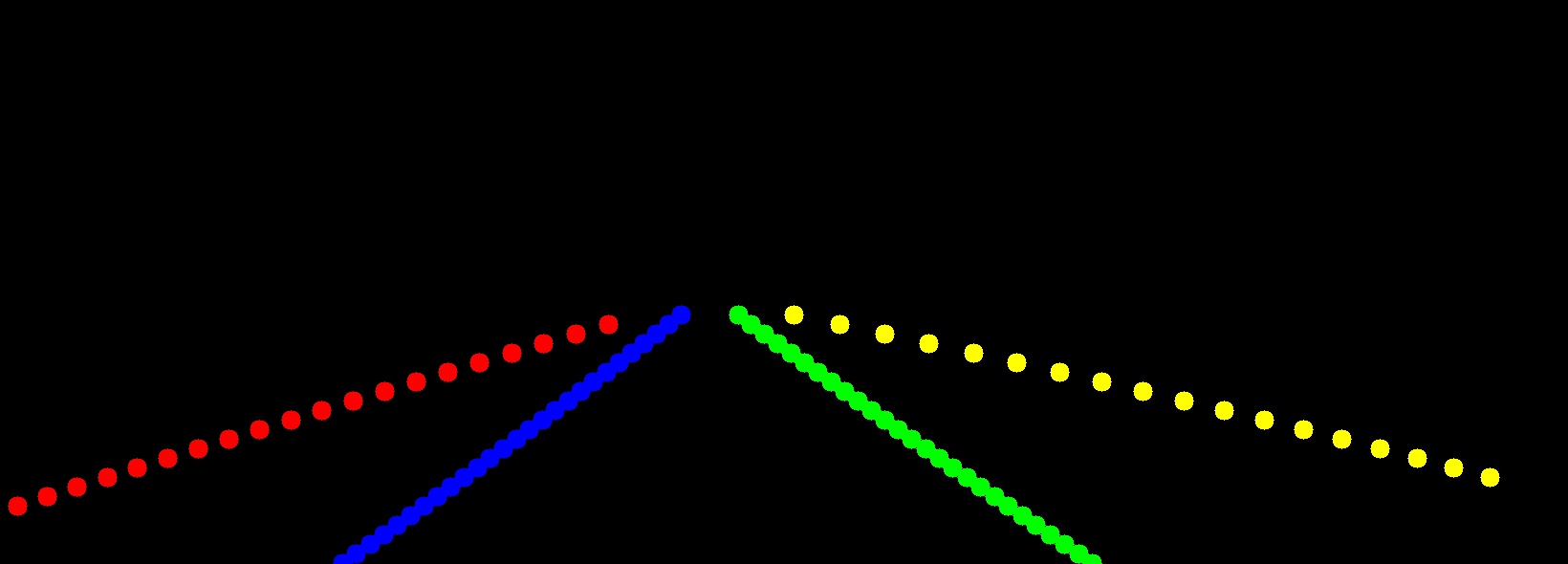}
     \end{subfigure}%
     \vspace{.3ex}
     
     \begin{subfigure}[b]{0.32\linewidth}
         \centering
         \includegraphics[width=0.98\linewidth]{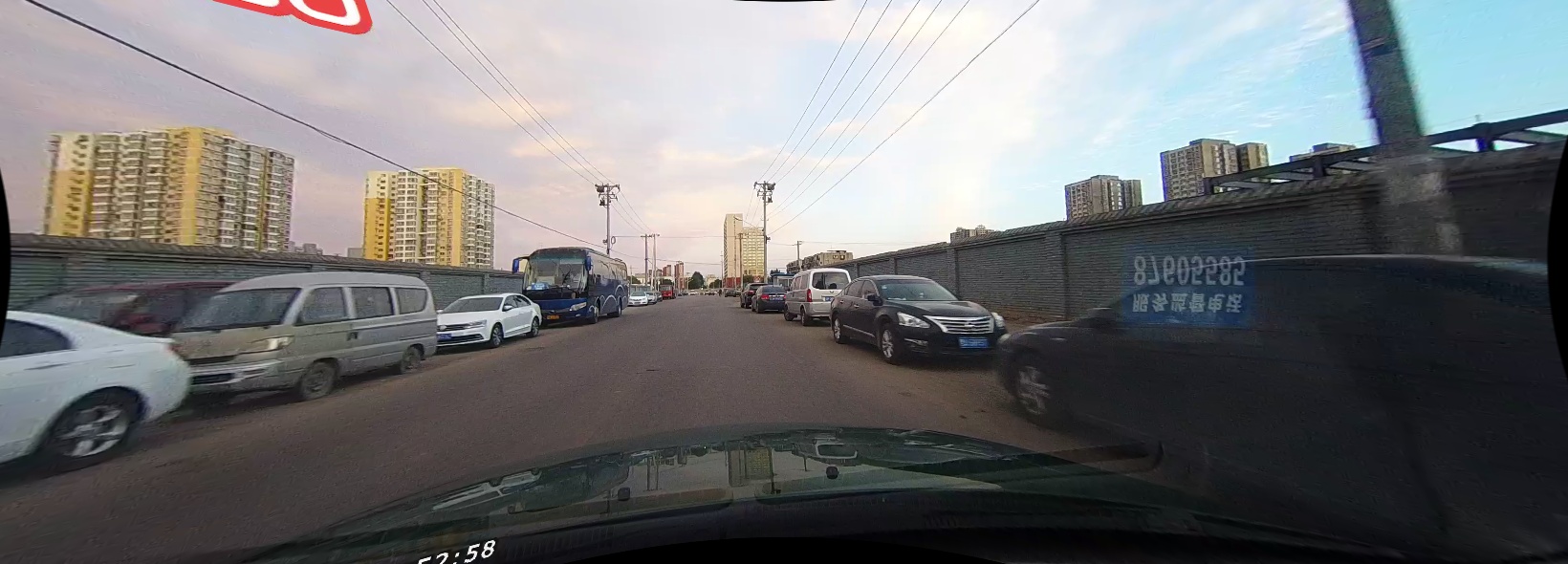}
     \end{subfigure}%
     \begin{subfigure}[b]{0.32\linewidth}
         \centering
         \includegraphics[width=0.98\linewidth]{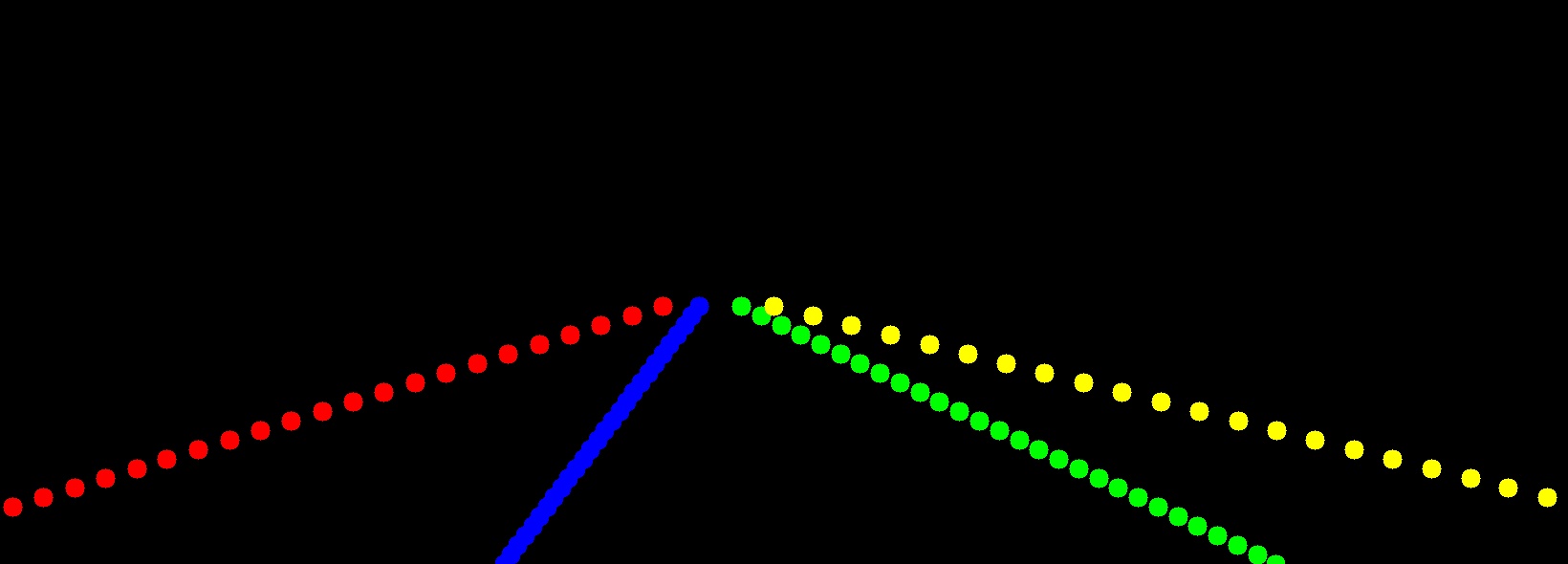}
     \end{subfigure}%
     \begin{subfigure}[b]{0.32\linewidth}
         \centering
         \includegraphics[width=0.98\linewidth]{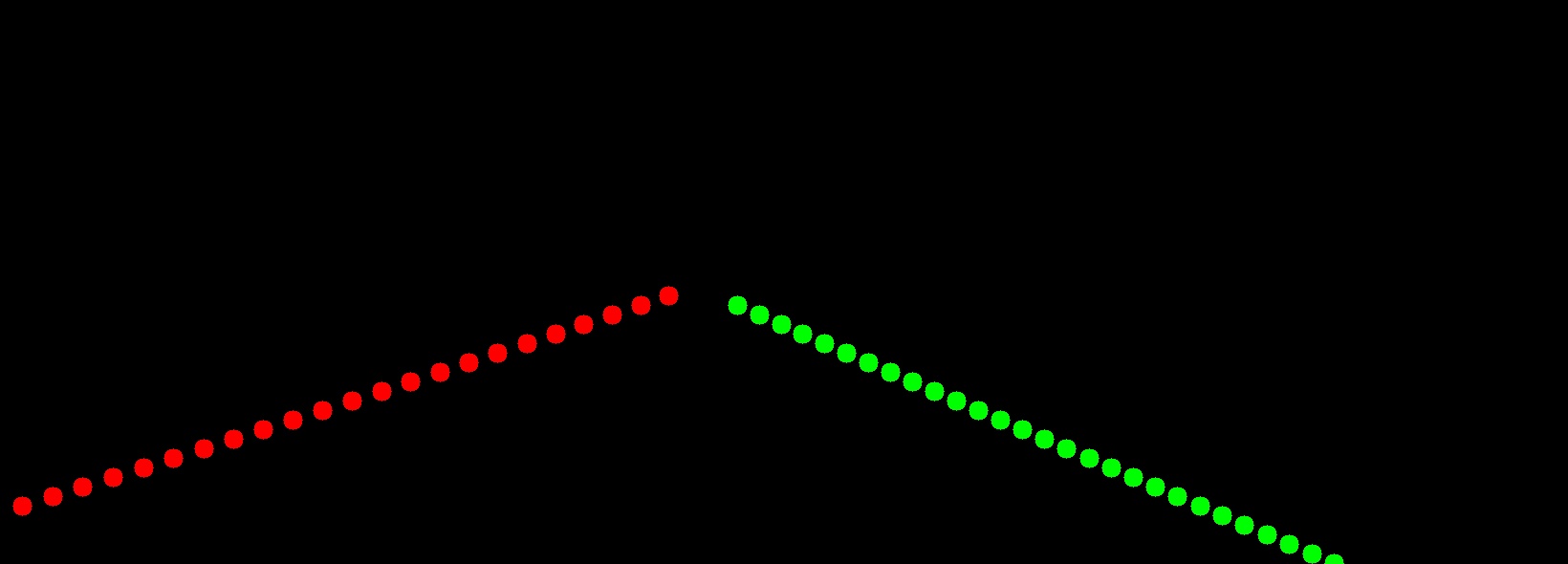}
     \end{subfigure}%
     \vspace{.3ex}
     
     \begin{subfigure}[b]{0.32\linewidth}
         \centering
         \includegraphics[width=0.98\linewidth]{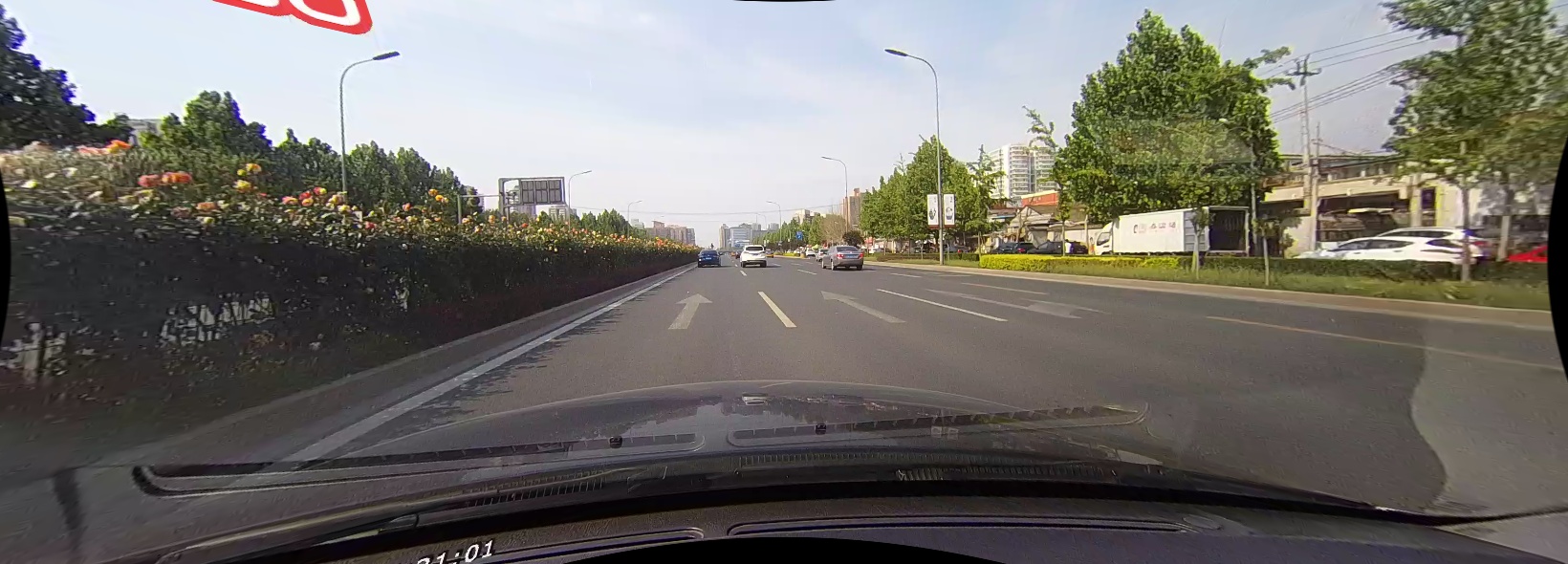}
     \end{subfigure}%
     \begin{subfigure}[b]{0.32\linewidth}
         \centering
         \includegraphics[width=0.98\linewidth]{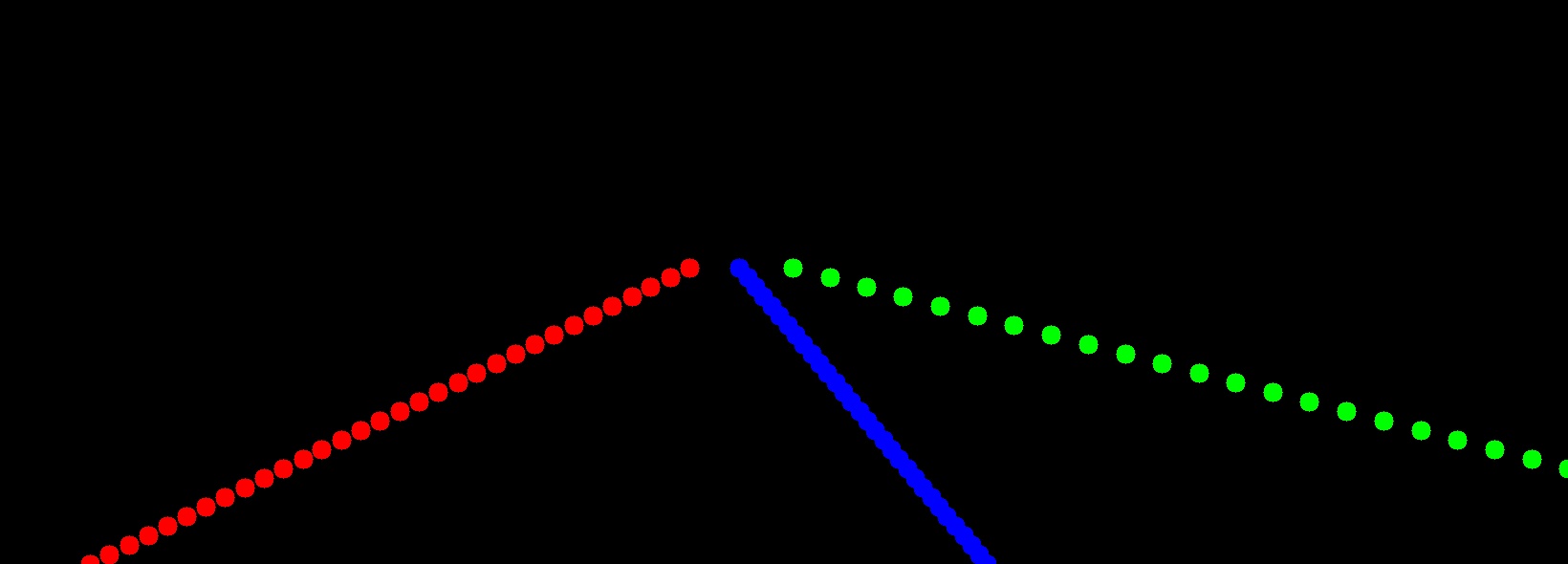}
     \end{subfigure}%
     \begin{subfigure}[b]{0.32\linewidth}
         \centering
         \includegraphics[width=0.98\linewidth]{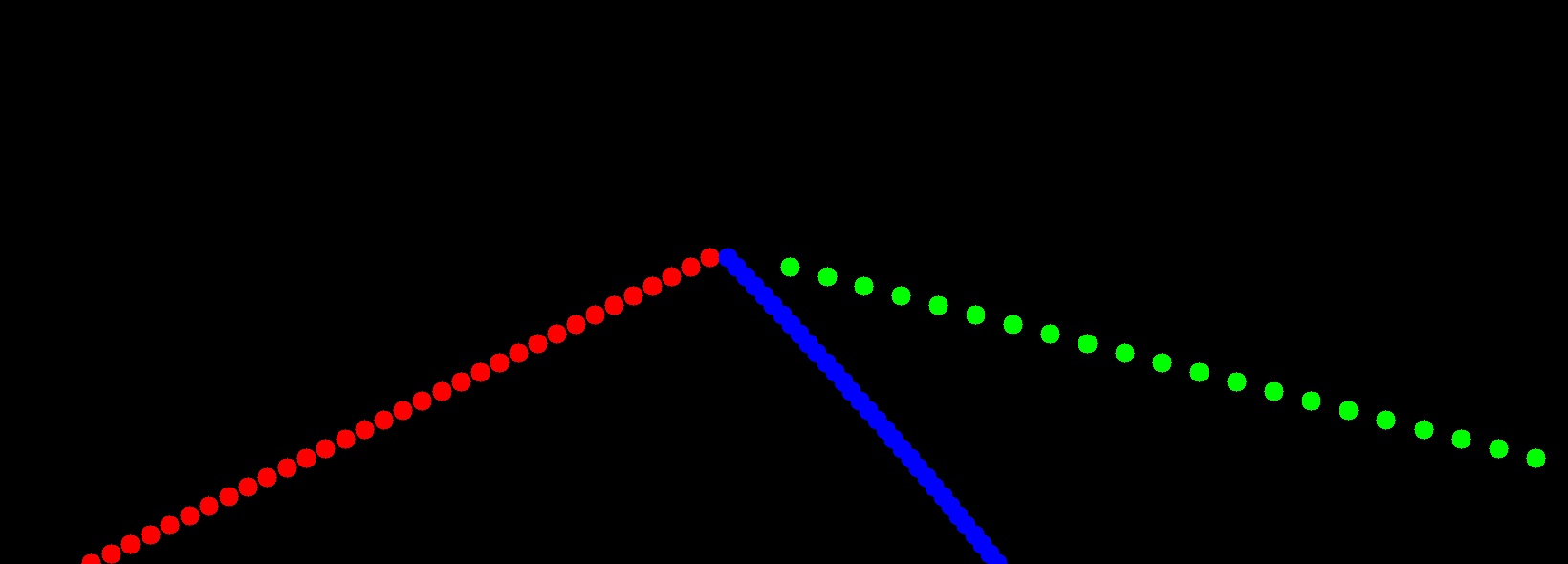}
     \end{subfigure}%
     \vspace{.3ex}

     \begin{subfigure}[b]{0.32\linewidth}
         \centering
         \includegraphics[width=0.98\linewidth]{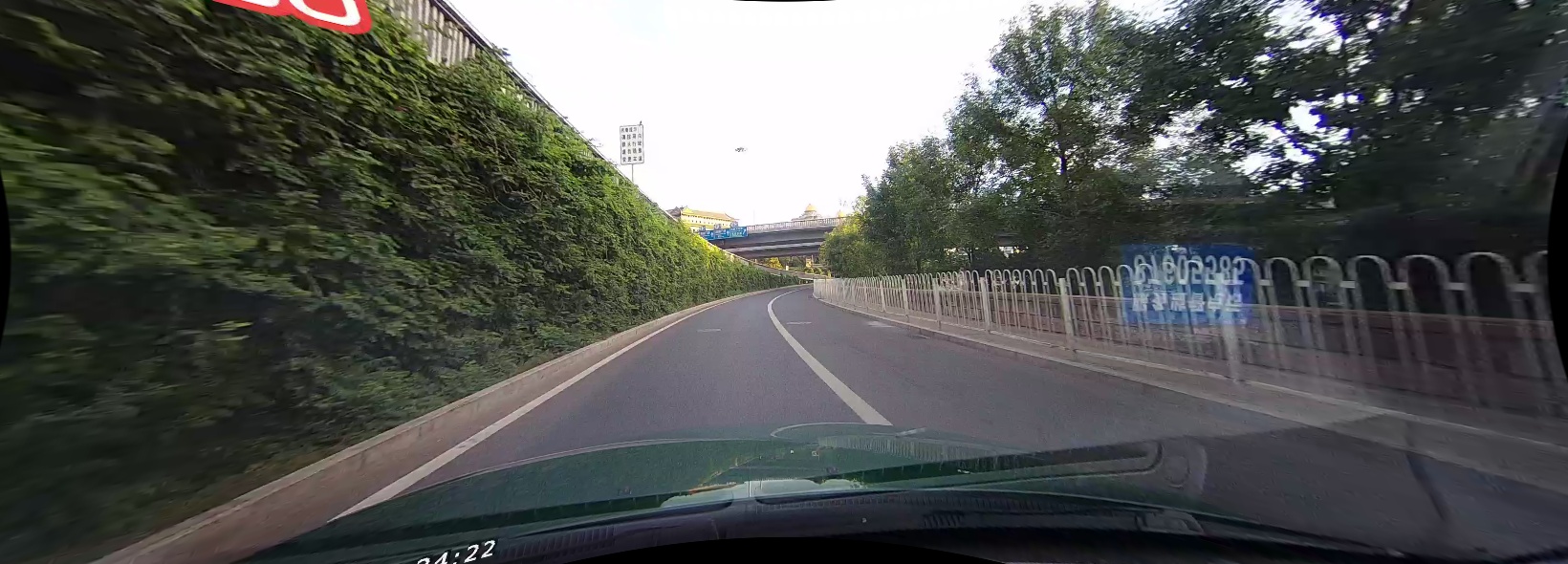}
     \end{subfigure}%
     \begin{subfigure}[b]{0.32\linewidth}
         \centering
         \includegraphics[width=0.98\linewidth]{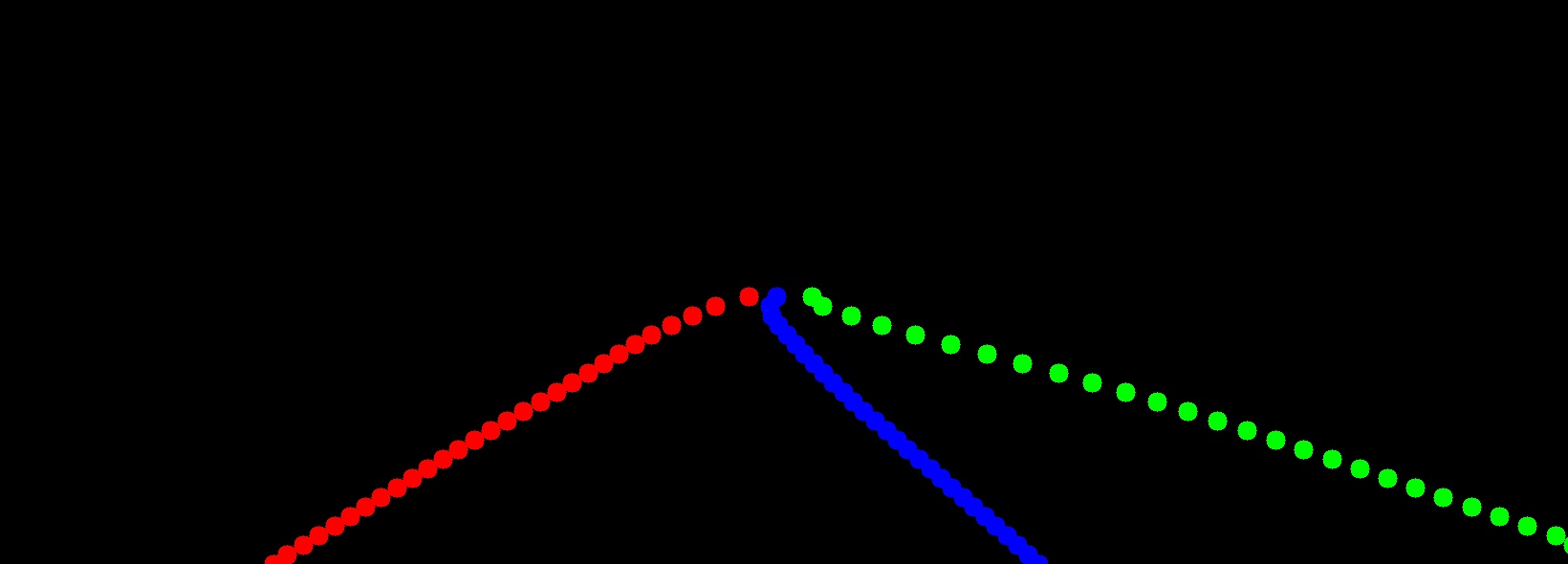}
     \end{subfigure}%
     \begin{subfigure}[b]{0.32\linewidth}
         \centering
         \includegraphics[width=0.98\linewidth]{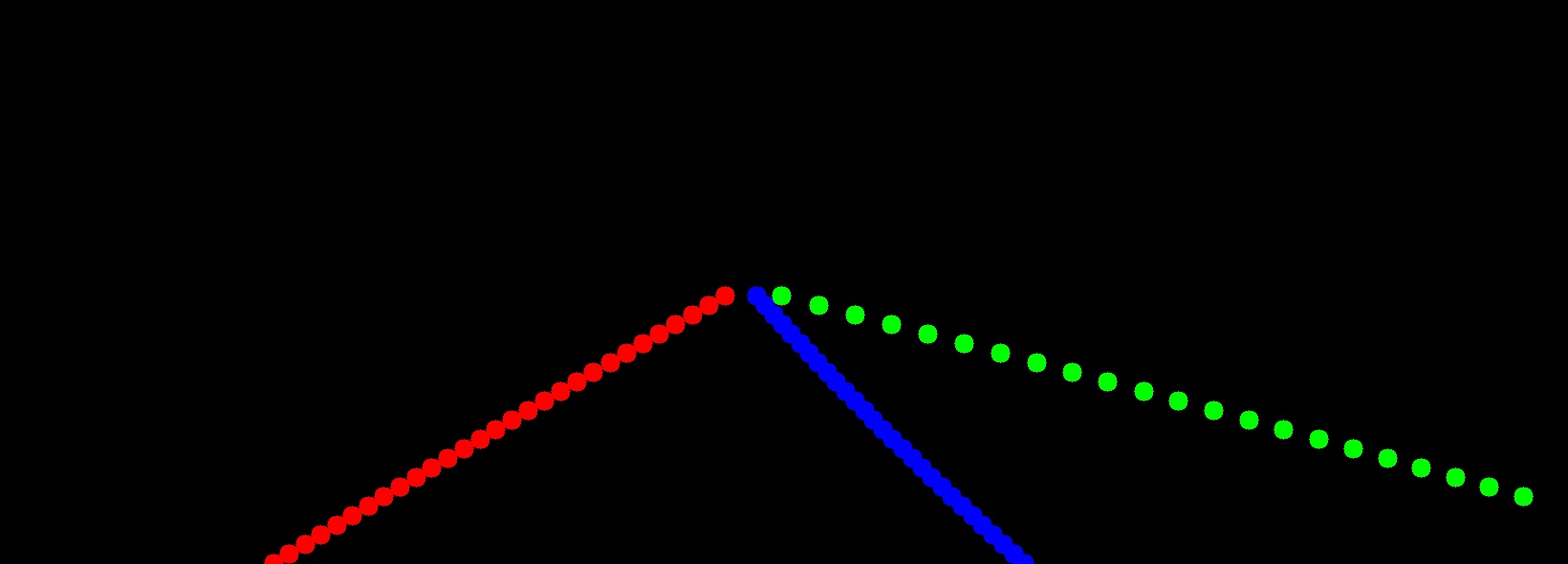}
     \end{subfigure}%
     \vspace{.3ex}

     \begin{subfigure}[b]{0.32\linewidth}
         \centering
         \includegraphics[width=0.98\linewidth]{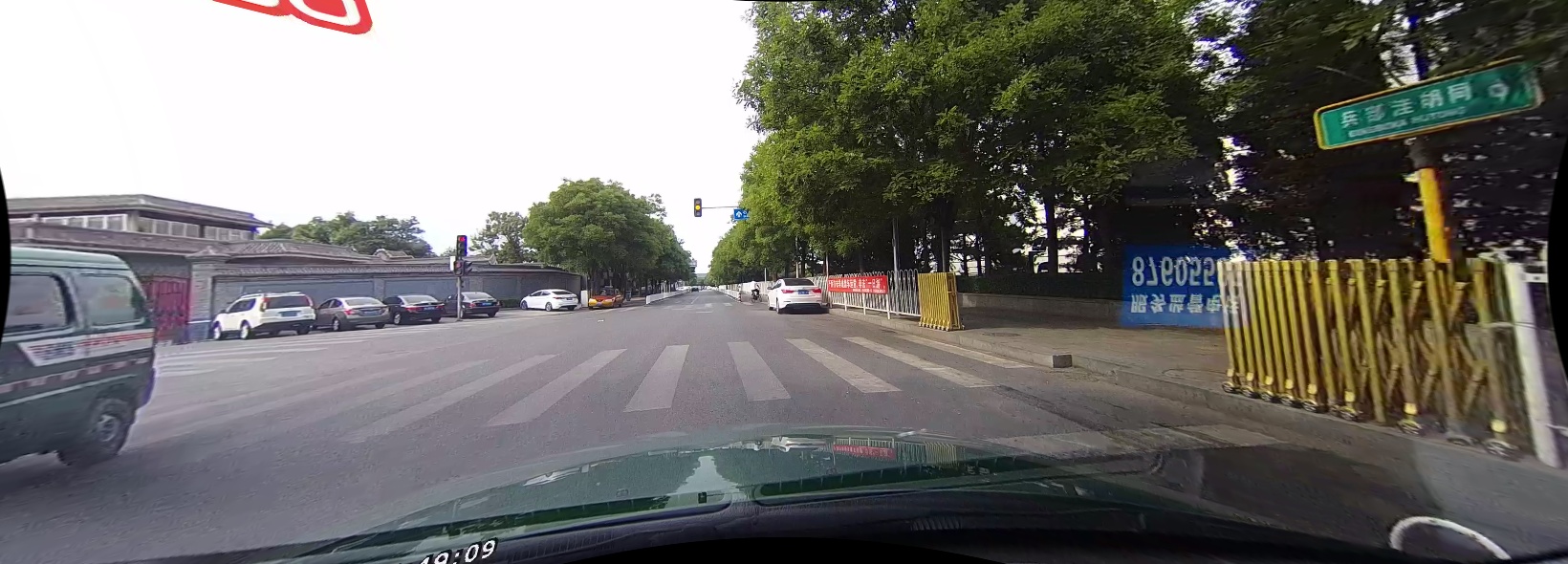}
     \end{subfigure}%
     \begin{subfigure}[b]{0.32\linewidth}
         \centering
         \includegraphics[width=0.98\linewidth]{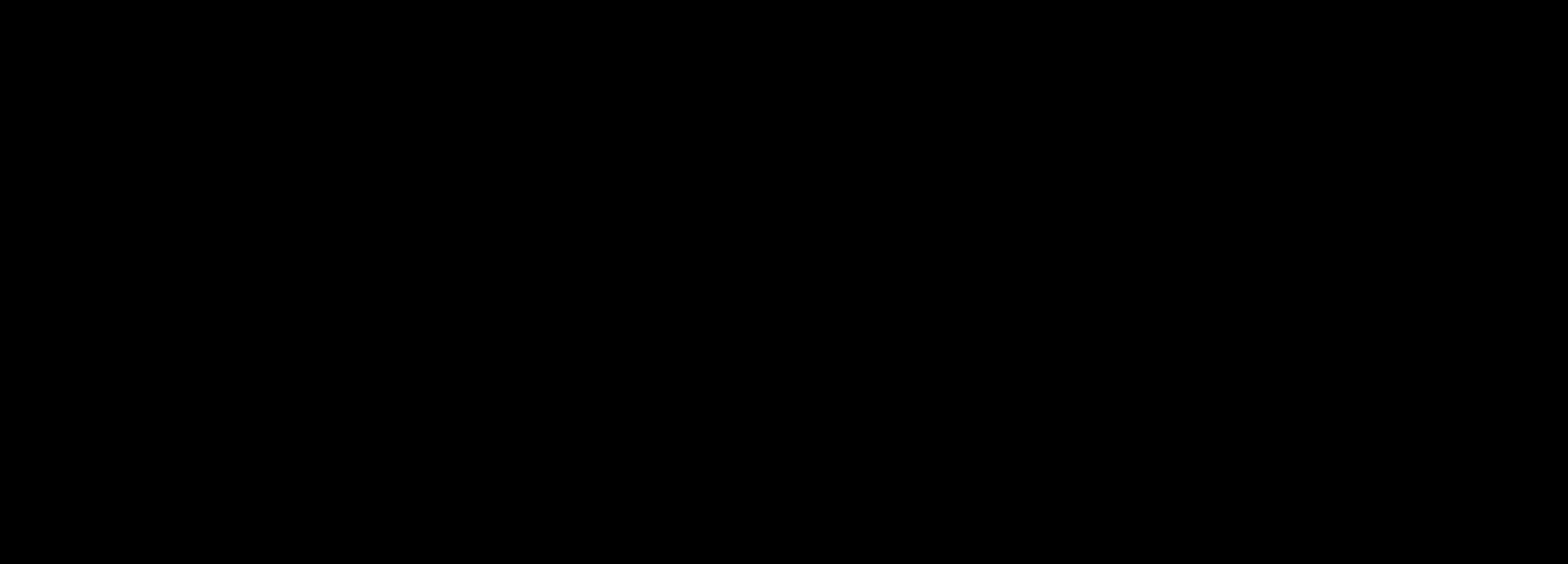}
     \end{subfigure}%
     \begin{subfigure}[b]{0.32\linewidth}
         \centering
         \includegraphics[width=0.98\linewidth]{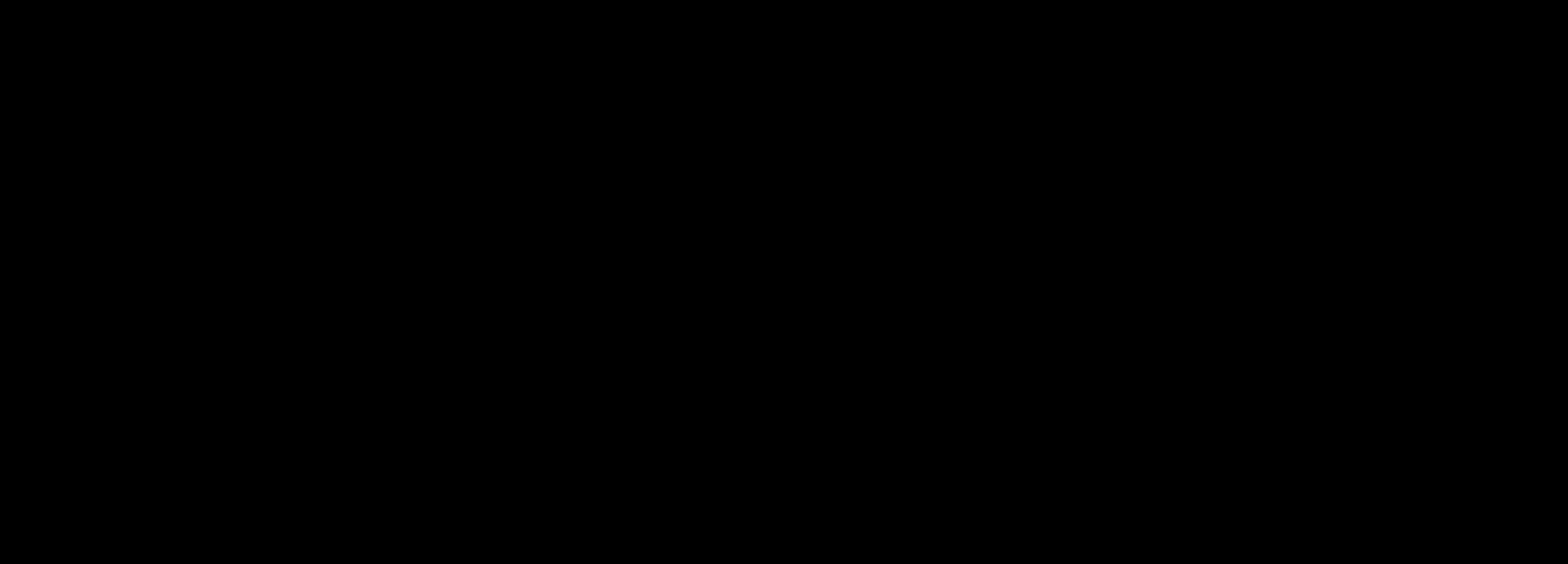}
     \end{subfigure}%
     \vspace{.3ex}
     
     \begin{subfigure}[b]{0.32\linewidth}
         \centering
         \includegraphics[width=0.98\linewidth]{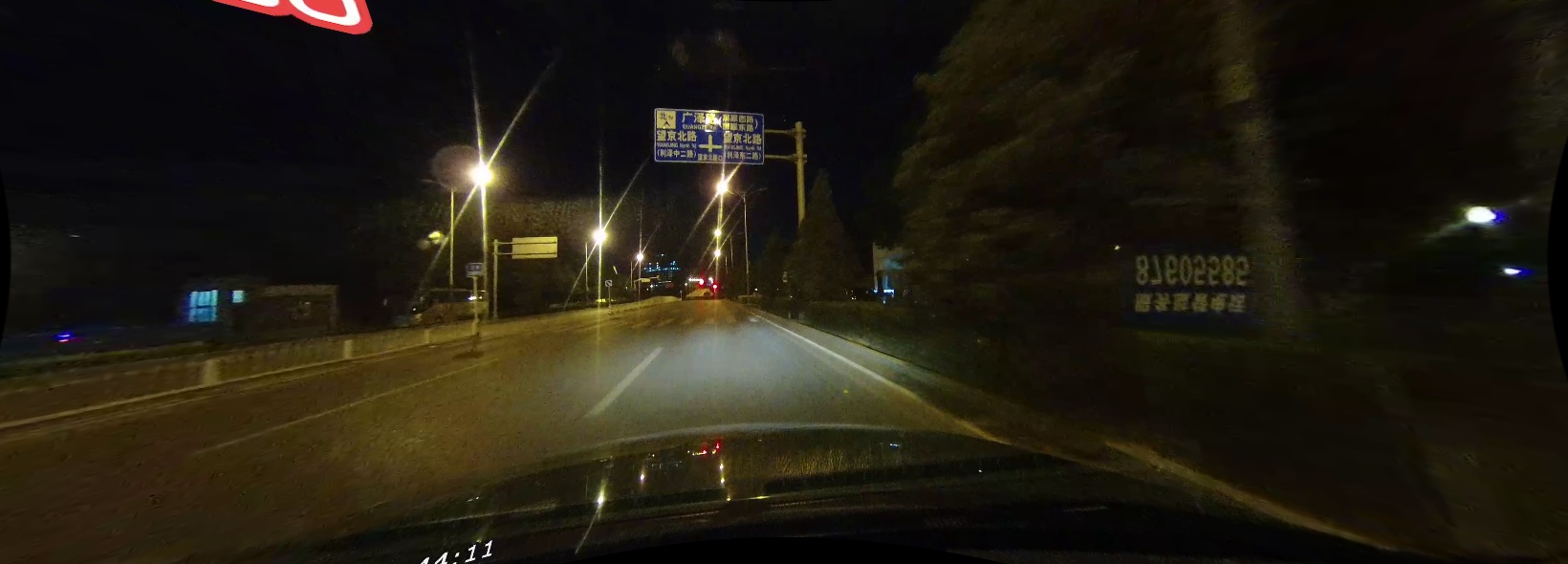}
         \caption{Input Image}
         \label{sfi:sf1}
     \end{subfigure}%
     \begin{subfigure}[b]{0.32\linewidth}
         \centering
         \includegraphics[width=0.98\linewidth]{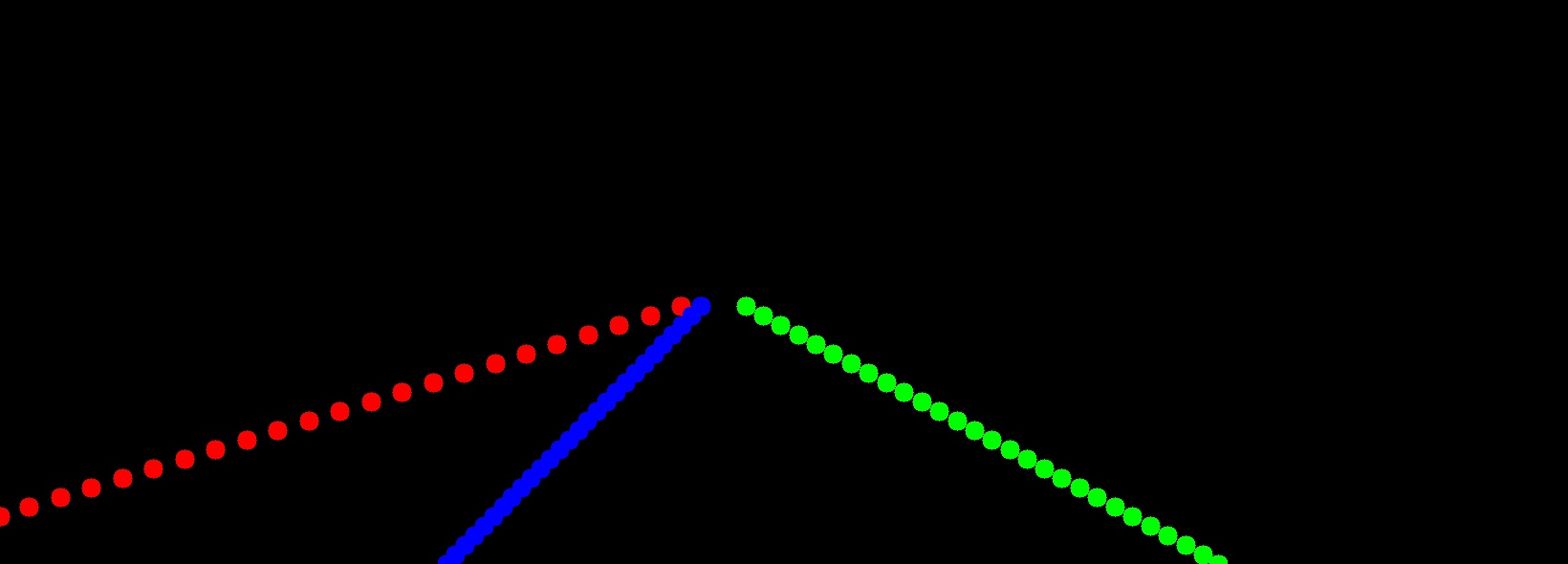}
         \caption{Ground Truth}
         \label{sfi:sf2}
     \end{subfigure}%
     \begin{subfigure}[b]{0.32\linewidth}
         \centering
         \includegraphics[width=0.98\linewidth]{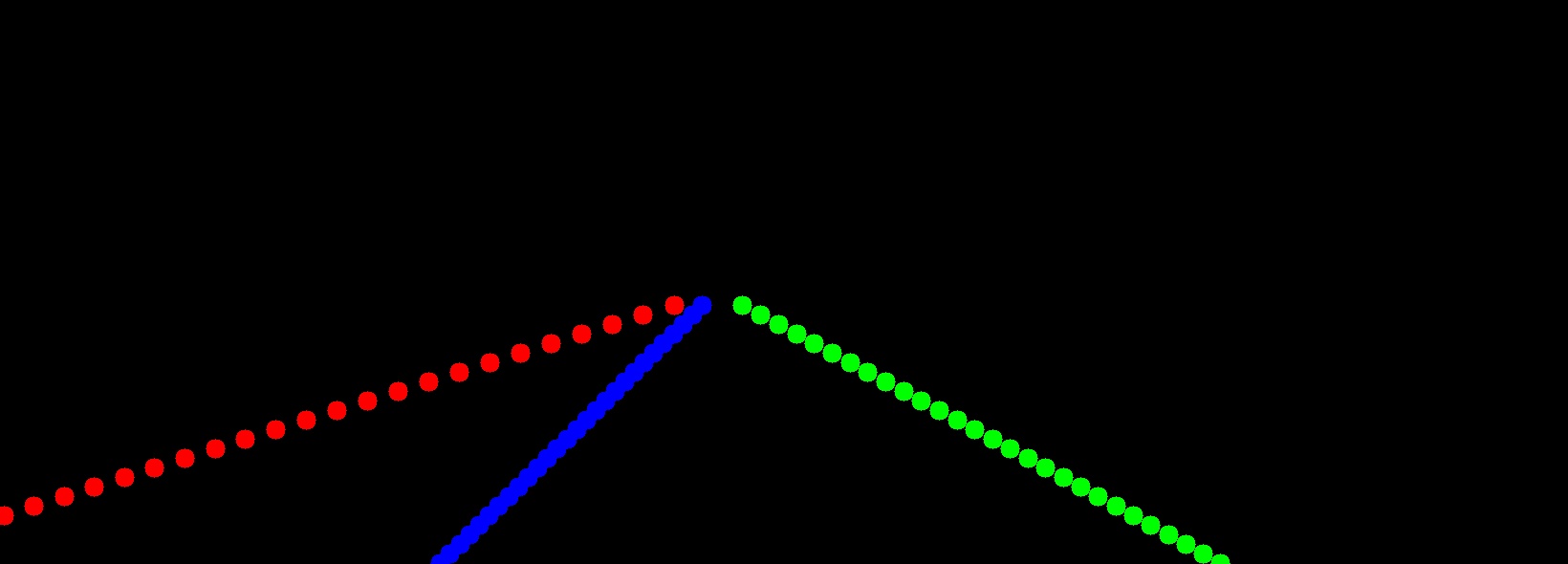}
         \caption{Prediction}
         \label{sfi:sf3}
     \end{subfigure}%
     \vspace{.3ex}
    
    \caption{ 
    Visualization of results on CULane. The nine rows represent the nine scenarios in CULane; Normal, Crowded, Dazzle light, Shadow, No line, Arrow, Curve, Crossroad and Night respectively.\\}
    
    \label{fi:pictorial_results}
    \vspace{-5ex}
\end{figure}

The performance of our method on the CULane benchmark dataset is compared against state-of-the-art lane detection approaches in Table \ref{ta:results}. The number of false positives are displayed under the ``Cross'' category since there are no true positives in the ground truth for that category. The inference speed is measured by taking the average frames per second (FPS) value for 1000 runs including the forward pass of the model and the post-processing steps. The number of multiply-accumulate operations in billions is represented in the ``GMACs'' column. For a fairer comparison, we measured the speed of \cite{qin2020ultra} under the same conditions as ours.

\begin{table}
\caption{Performance on the embedded system}
\vspace{-1ex}
\label{ta:opt}
\begin{center}
\normalsize
\begin{tabular}{|l|c|c|}
\hline
\textbf{Model} & \textbf{F1-measure} & \textbf{Speed (FPS)} \\
\hline
Pytorch Model & 74.02 & 23   \\
\hline
TensorRT Engine (FP32) & 74.02 & 35  \\
\hline
TensorRT Engine (FP16) & 74.03 & 56  \\
\hline

\end{tabular}
\end{center}
\vspace{-4ex}
\end{table}

It can be observed that while being the fastest, our method achieves competitive results with other state-of-the-art methods in F1-measure. Our method also uses the least number of multiply-accumulate operations (MACs) which highlights the efficiency of our formulation. The low number of false positives in the ``Cross'' category validates the effectiveness of our false positive suppression technique. Compared to the segmentation based methods \cite{pan2018SCNN, hou2019learning}, the inference speed improves substantially while providing better results at the same time. When compared with \cite{qin2020ultra}, which is the fastest among other approaches, our method achieves better results with a 6.6\% increase in F1-measure. While we obtain comparable performance with \cite{CurveLane-NAS} and \cite{yoo2020end}, a direct comparison cannot be made in terms of the speed, as their inference speeds are not mentioned. Although \cite{pinet_2021}, \cite{resa_2020}, \cite{tabelini2021keep} and \cite{FOLO_2021_CVPR} achieves on par or better results than our method, the low inference speeds of their best performing models act as a barrier for real-time implementation especially on resource constrained environments.

The performance of the Pytorch model and the generated FP32 and FP16 TensorRT engines on the Nvidia Jetson AGX Xavier are shown in Table \ref{ta:opt} in terms of the F1-measure and speed. The inference speed is calculated as the average frames per second value for inferencing a locally captured video within the ROS ecosystem. It can be observed that while the accuracy stays almost the same, the inference speed has increased significantly by optimizing and quantizing the model through TensorRT. 

Qualitative results obtained by our lane detector model are visualized in Fig. \ref{fi:pictorial_results} for the nine categories in the CULane dataset. In addition, locally captured street view images that encompass a range of road scenarios including urban, rural and expressway conditions are inferenced in order to assess the robustness of our trained model. Some of those results are shown in Fig. \ref{fi:local_lane_qualitative_results}.



\begin{table}[t]
\caption{Ablation study results on CULane}
\vspace{-1ex}
\label{ta:Ablation}
\begin{center}
\normalsize
 \begin{tabular}{|@{}l|c|c|}  
 \hline
 \textbf{Proposed Method} & \textbf{F1-measure} & \textbf{Speed (FPS)}\\
 \hline
Base Model & 71.25 & 502 \\
 \hline
 + FP Suppression (length) & 72.76 & 489\\
 \hline
+ FP Suppression (linearity) & 73.90 & 447\\
 \hline
+ Curve Fitting & 74.03 & 411 \\
 \hline
\end{tabular}
\end{center}
\vspace{-3ex}
\end{table}

\subsection{Ablation Study}
\label{ssec:ablation}

As an ablation study, each of the proposed methods is evaluated in terms of the speed and the F1-measure, as given in Table \ref{ta:Ablation}. The first line contains the results of the base model, and the FPS value is calculated based on the forward inference time on the GPU. The high FPS value shows the efficiency of our proposed light-weight network architecture with reduced multiply-accumulate operations (MACs). The rest of the lines show how the proposed post-processing techniques contribute towards increasing the accuracy. However, employment of each method reduces the FPS value, especially because these algorithms run on the CPU.

\section{Conclusion}
\label{sec:conc}
In this work, we proposed a simple, light-weight, end-to-end deep learning based network architecture coupled with the row-wise classification formulation for fast and efficient lane detection. Furthermore, we introduced a false positive suppression algorithm based on the length of the lane segment and the Pearson correlation coefficient, and a second-order polynomial fitting method as post-processing techniques. Collectively, our approach surpasses state-of-the-art with regard to speed reaching up to 411 FPS, while achieving competitive results in terms of accuracy, as justified in the qualitative and quantitative experiments carried out on the CULane benchmark dataset. We further demonstrated the capability of our light-weight network architecture to perform in real-time, by optimizing and quantizing our trained model using TensorRT and deploying on an embedded system while integrating with ROS, which achieves a high inference speed of 56 FPS. The inference results for the locally captured street view images show how well our method generalizes for the task of lane detection.




{
\bibliographystyle{ieeetran}
\bibliography{egbib}
}

\end{document}